\documentclass[acmsmall,screen,nonacm]{acmart}

\usepackage{multirow}
\usepackage{booktabs}
\usepackage{algorithm}
\usepackage{algpseudocode}
\usepackage{xcolor}

\begin{document}

\title[Automatic Construction of Persian E-Commerce Dialogue Datasets]{Automatic Construction of Persian E-Commerce Dialogue Datasets for Conversational Product Search}

\author{Fariba Nafar}
\email{fariba\_nafar@iust.ac.ir}
\affiliation{%
  \institution{Iran University of Science and Technology}
  \department{School of Computer Engineering}
  \city{Tehran}
  \country{Iran}}

\author{Hassan Naderi}
\authornote{Corresponding author.}
\email{naderi@iust.ac.ir}
\affiliation{%
  \institution{Iran University of Science and Technology}
  \department{School of Computer Engineering}
  \streetaddress{Hengam St}
  \city{Tehran}
  \postcode{16846-13114}
  \country{Iran}}

\renewcommand{\shortauthors}{Nafar and Naderi}

\begin{abstract}
Conversational product search lets users find products through a multi-turn dialogue instead of a single keyword query, but Persian lacks the large conversational datasets needed to train and evaluate such systems; existing Persian e-commerce dialogue resources cover few categories and small catalogs. We present a framework that builds such a dataset automatically from a real e-commerce catalog and release the resulting dataset. The framework crawls 12,926 products in seven categories of the Digikala marketplace, groups them into 170 semantically coherent clusters, and selects the question the assistant asks at every turn with the Regularized Partition Balance (RPB) score. The RPB score accounts for feature coverage, the probability that the user can answer and the number of values, and generalizes entropy-based question selection. An LLM then verbalizes each question plan into a natural Persian dialogue. In a simulation on the full catalog, the RPB score needs 0.11 fewer turns than the strongest entropy baseline when the difficulty of answering differs between features, and is practically equivalent to it when it does not. Using only a free LLM API, we generated 1,000 requirement-driven dialogues covering all 170 clusters, with turn-level feature annotations; automatic checks and an LLM judge confirm their consistency with the catalog, while their realism is lower than that of the first, GPT-4o-generated version. Code and data are publicly available.
\end{abstract}



\maketitle

\section{Introduction}\label{sec:intro}
Product search in e-commerce is shifting from static keyword queries to multi-turn conversations, in which the system asks clarifying questions and the user progressively reveals what they need \citep{radlinski2017theoretical,zhang2018towards,zamani2023conversational}. Conversational Product Search (CPS), strengthened by Large Language Models (LLMs), can bridge the gap between the user's intent and the product representation and thereby improve search efficiency and user satisfaction. Building robust CPS systems, however, requires large, high-quality, task-specific datasets. For low-resource languages such as Persian, the lack of such datasets is a major bottleneck, compounded by the language's morphological and orthographic complexity \citep{khashabi2021parsinlu,farahani2021parsbert}.

A related obstacle is the lack of culturally relevant and clearly structured product catalogs. Recent frameworks for English CPS use extensive, high-quality catalogs to generate realistic training dialogues \citep{li2025wizardshoppingtargetorientedecommerce,wang2025shoppingbenchrealworldintentgroundedshopping}, but no comparable resource exists for the Persian market, and the effectiveness of these frameworks depends on a catalog that covers many categories and reflects the attributes, terminology and products of the target market.

Existing generation frameworks select the product features that the dialogue discusses either randomly or by how well they separate related products. Neither handles a property of real catalogs: many features are recorded for only some products, some are unique to a single item, and some cannot realistically be answered by a user. Dialogues that ask about such features are unnatural, and excluding them manually is time-consuming. Real users instead give feature information incrementally, in a way that narrows down the relevant products.

This paper makes the following contributions:
\begin{itemize}
    \item A framework that builds a Persian CPS dataset automatically from a real catalog: crawling, semantic clustering, question planning, contextual augmentation and LLM-based verbalization (Section~\ref{sec:framework}).
    \item The RPB score for choosing the next question, which accounts for coverage, answerability and the number of values and generalizes information gain (Section~\ref{subsec:feature_metric}).
    \item An evaluation of the score with a simulated user on the full Digikala catalog against random selection and two entropy variants (Section~\ref{subsec:rpb-eval}).
    \item A comparison of three LLM generation strategies, RPB-planned versus random features and three free LLMs, followed by an LLM-judge and human assessment (Sections~\ref{subsec:method-selection} and~\ref{sec:human}).
    \item A dataset of 1,000 Persian requirement-driven dialogues with turn-level feature annotations, generated with a free LLM API so that it can be reproduced and extended at no cost, together with the code (Section~\ref{sec:dataset}).
\end{itemize}

\section{Related work}\label{sec:related}

\subsection{Conversational search and recommendation}
Conversational search was formalized as a mixed-initiative interaction in which the system may ask questions to resolve the user's need \citep{radlinski2017theoretical}; \citet{zamani2023conversational} survey the area. In product search, \citet{zhang2018towards} proposed the ``system ask, user respond'' paradigm, in which the system asks about product aspects, and \citet{bi2019conversational} used negative feedback on presented items; negative feedback was later also used to choose clarifying questions \citep{bi2021asking}. Which product-aspect question to ask has been learned with representation learning \citep{zou2022learning} and deep reinforcement learning \citep{montazeralghaem2022learning}, which require training dialogues, whereas the RPB score is computed directly from the catalog. MG-ShopDial \citep{bernard2023mgshopdial} is a crowd-sourced English dialogue dataset in which shoppers pursue several goals in one conversation. Asking clarifying questions has been studied with the Qulac and ClariQ collections \citep{aliannejadi2019asking,aliannejadi2021building}. Conversational recommender systems decide which attribute to ask about and when to recommend \citep{christakopoulou2016towards,sun2018conversational,lei2020estimation,lei2020interactive}; see \citep{jannach2021survey,gao2021advances} for surveys. \citet{zou2020towards} select the question that best splits the candidate items, which is closest to our setting, and Uncertainty of Thoughts \citep{hu2024uncertainty} lets an LLM choose questions by their expected information gain over a set of candidates. Conversational recommenders are commonly evaluated with simulated users \citep{zhang2020evaluating}, increasingly with LLM-based ones \citep{wang2023rethinking}; such simulators, however, may leak the target or behave unlike real users \citep{yoon2024evaluating,zhu2024reliable}, which is why our simulated user answers from the catalog values of the target. Entropy, or information gain, is the classical criterion for choosing such questions \citep{quinlan1986induction}; the RPB score extends it with the coverage, answerability and cardinality of real catalog features.

\subsection{Dialogue datasets and LLM-based generation}
Crowd-sourced conversational recommendation datasets such as ReDial \citep{li2018towards} are expensive to build. LLMs make it possible to synthesize dialogues at scale \citep{kim2023soda,suresh-etal-2025-diasynth}, either with a single prompt that produces the whole conversation, or with two LLM agents that play the participants \citep{duan-etal-2024-botchat,de-baer-etal-2025-single}, or by rewriting templates \citep{shim2025tooldialmultiturndialoguegeneration}; \citet{long2024synthetic} survey the generation, curation and evaluation of such synthetic data. PLACES \citep{chen2023places} synthesizes social conversations by few-shot prompting, and several datasets for conversational recommendation were built with LLMs: LLM-REDIAL \citep{liang2024llmredial} turns real user behaviour into dialogues, Pearl \citep{kim2024pearl} grounds simulated users in personas and reviews, and Talk the Walk \citep{leszczynski2023talk} first plans a sequence of item sets over a music catalog and then verbalizes it, a plan-then-verbalize design similar to ours. Two recent frameworks generate e-commerce dialogues. ShoppingBench \citep{wang2025shoppingbenchrealworldintentgroundedshopping} samples real products for given intents (e.g., ``Knowledge'' or ``Coupon \& Budget''), extracts their titles and attributes and uses GPT-4 to simulate user queries, yielding four shopping intents of increasing difficulty. TRACER \citep{li2025wizardshoppingtargetorientedecommerce} simulates conversations between customer and seller agents: it samples the customer's preferences, plans the search trajectory with a decision tree that narrows down the products, and verbalizes it with an LLM, yielding the Wizard of Shopping dataset. The first framework selects features randomly; the second chooses those that best separate related products, which, as we show, is not optimal when features have low coverage. Generated dialogues are increasingly evaluated with an LLM as a judge \citep{zheng2023judging,liu2023geval}, which correlates with human ratings but tends to favour outputs of the same model \citep{panickssery2024llm} and is sensitive to the order of the compared outputs \citep{wang2024fair}; see \citet{gu2024survey} for a survey.

\subsection{Persian e-commerce dialogue datasets}
Persian NLP benchmarks have grown \citep{khashabi2021parsinlu,farahani2021parsbert}, including benchmarks of LLMs \citep{abaskohi2024benchmarking,ghahroodi2024khayyam} and of text embeddings \citep{zinvandi2025famteb}, but resources for shopping dialogues remain small. Large real-catalog resources such as the Shopping Queries dataset \citep{reddy2022shopping} exist for English, Spanish and Japanese, but not for Persian. \citet{9345639} addressed data scarcity for Persian online-shopping dialogue systems with two generation methods: a fully simulated approach based on rules and templates, and a semi-simulated one that used human translation to improve naturalness and diversity; combining them improved the natural-language-understanding module. PerSHOP \citep{mahmoudi2024pershoppersiandataset} was collected through crowd-sourcing with a web application. MegaChat \citep{rahmani2025megachat} is a synthetic Persian question-answer dataset for evaluating sales chatbots, with single questions rather than multi-turn product search, and HamRaz \citep{abbasi2025hamraz} builds Persian dialogues with LLM agents in the therapy domain. Both are limited to a few categories, small catalogs and products with few attributes, which yields low-variety dialogues. Our method applies to all categories of a large catalog and generates diverse dialogues grounded in real product data.

\section{Dialogue dataset constructor}\label{sec:framework}
\begin{figure}[htbp]
    \centering
    \includegraphics[width=\textwidth]{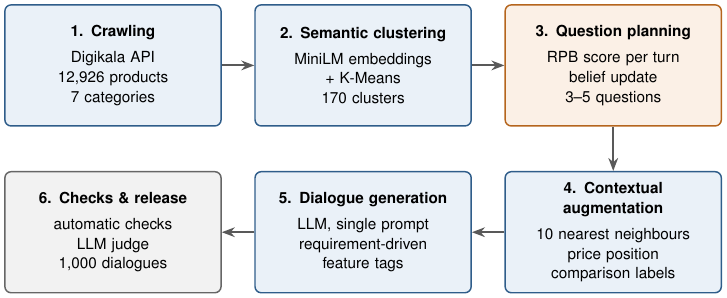}
    \caption{Framework for dialogue dataset construction. The question planning step (highlighted) uses the RPB score of Section~\ref{subsec:feature_metric}.}
    \label{fig:framework}
\end{figure}

Figure~\ref{fig:framework} outlines the pipeline. Product data are collected through the Digikala API, products are clustered by their titles and descriptions, the RPB score plans the questions of each dialogue within the target's cluster, each product is augmented with comparisons to its nearest neighbours, and an LLM verbalizes the plan into a dialogue.

\subsection{Product data crawling}
The product data were collected through the API of Digikala, the largest Iranian e-commerce platform \citep{digikala}. We selected seven categories to ensure diversity: smartphones, laptops, stationery, home decoration, carpets, wearable gadgets and books. For each product, the API returns its brand, category, price, rating, description and technical specifications as a nested JSON object. Table~\ref{tab:data-collection} gives the number of products per category.

\begin{table}[h]
\centering
\begin{tabular}{lr}
\toprule
\textbf{Category} & \textbf{Products}\\
\midrule
Smartphones & 1,555 \\
Laptops & 1,715 \\
Stationery & 1,998 \\
Home decoration & 2,000 \\
Wearable gadgets & 1,878 \\
Carpets & 1,781 \\
Books & 1,999 \\
\midrule
Total & 12,926 \\
\bottomrule
\end{tabular}
\caption{Distribution of the collected products by category.}
\label{tab:data-collection}
\end{table}

\subsection{Semantic clustering of products}
Questions are only meaningful among comparable products, so products are first grouped by semantic similarity using their titles and descriptions. We compared TF-IDF \citep{salton1988term} with the multilingual MiniLM sentence encoder \texttt{paraphrase-multilingual-MiniLM-L12-v2} \citep{10.5555/3495724.3496209,reimers2019sentence,reimers2020making} as text representations, and K-Means \citep{macqueen1967some}, hierarchical clustering, Gaussian mixture models and DBSCAN \citep{ester1996density} as clustering algorithms, using the Silhouette score \citep{rousseeuw1987silhouettes}, the Davies--Bouldin index \citep{davies1979cluster} and the Calinski--Harabasz index \citep{calinski1974dendrite}. To validate the configuration, we first checked whether it recovers the seven original categories for cluster counts between 3 and 20, and then applied it with finer granularity (Section~\ref{subsec:clustering}).

\subsection{Regularized Partition Balance (RPB) score}
\label{subsec:feature_metric}

To minimize the number of turns needed to find a product, the assistant should ask, at every turn, the question that rules out the most irrelevant products. This section introduces a score that quantifies the discriminative power of a feature. Besides how the products are divided among the values of the feature, the score accounts for three properties of real e-commerce catalogs: (i) not every product has every feature, and products lacking a feature fall into a \emph{default group}; (ii) users may not know, or not care about, the answer to some questions, especially those about technical features; and (iii) questions about features with very many values are hard to answer and error-prone.

\paragraph{Notation.}
Let $S=\{s_1,\dots,s_n\}$ be the products of a cluster and $b_t(s)$ the system's belief that $s$ is the user's target after $t$ dialogue turns, with $\sum_{s\in S} b_t(s)=1$. In dynamic use, $b_0$ is proportional to product popularity (e.g., the number of ratings); in static use it is uniform, $b_0(s)=1/n$. For a feature $P$ with real values $\{1,\dots,m\}$, let $S_{P,i}$ be the products with value $i$ and $S_P^{\varnothing}$ the default group of products lacking $P$. The default group is not a value, since a user cannot give it as an answer. We define
\begin{equation}
\kappa_P = \sum_{s\notin S_P^{\varnothing}} b_t(s), \qquad
p_i = \frac{1}{\kappa_P}\sum_{s\in S_{P,i}} b_t(s), \qquad
N_P = \frac{\kappa_P^{2}}{\sum_{s\notin S_P^{\varnothing}} b_t(s)^{2}},
\end{equation}
where $\kappa_P$ is the weighted coverage of the feature, $p_i$ the share of value $i$ among the products that have the feature, and $N_P$ the effective number of candidates that have the feature (Kish's effective sample size \citep{kish1965survey}); under a uniform belief, $N_P$ equals the number of products that have the feature.

\paragraph{Definition.}
The RPB score of feature $P$ is
\begin{equation}\label{eq:rpb}
\mathrm{RPB}(P\mid b_t) = \bigl(1-\hat{\delta}_P\bigr)\,\kappa_P\,\Bigl[\,H(P) - \alpha\,\max\bigl(0,\ \log_2 m_P - \tfrac12\log_2 N_P\bigr)\Bigr]_{+},
\end{equation}
where $[x]_+=\max(0,x)$, $m_P$ is the number of values with a share of at least 1\% among the current candidates (at least two), and
\begin{equation}\label{eq:rpb-balance}
H(P) = -\sum_{i=1}^{m} p_i\log_2 p_i = \log_2 m - D_{\mathrm{KL}}\!\left(p \,\middle\|\, U_m\right),
\end{equation}
with $U_m$ the uniform distribution over $m$ values. The score has four components:
\begin{itemize}
    \item \textbf{Partition balance} $H(P)$. By Eq.~\eqref{eq:rpb-balance}, the term $\log_2 m$ rewards the number of values and the term $D_{\mathrm{KL}}$ penalizes the distance of the partition from perfect balance. In bits, $H(P)$ is the information that the user's answer provides about the target product.
    \item \textbf{Coverage} $\kappa_P$. The answer is informative only about products that have the feature; a feature recorded for a small fraction of the products narrows down only that fraction of the search space, however balanced it is.
    \item \textbf{Answerability} $1-\hat{\delta}_P$, where $\hat{\delta}_P$ is the probability of a ``don't know'' or ``don't care'' answer. The product $(1-\hat{\delta}_P)\,\kappa_P$ is the probability that the answer is usable, so the score approximates the expected usable information of a question, in bits.
    \item \textbf{Cardinality penalty}. Features with more than about $\sqrt{N_P}$ values are penalized: two balanced questions with $\sqrt{N_P}$ values each suffice to identify one of $N_P$ candidates, and more values make the question harder and the answer more error-prone. The penalty is logarithmic and dimensionless, so $\alpha$ has the same meaning in clusters of any size, and it is one-sided because features with few values are already penalized through a small $H(P)$.
\end{itemize}
The score has several useful properties. First, a feature with a single real value among the products that have it scores zero, even if it has a large default group. Second, in the ideal case ($\kappa_P=1$, $\hat{\delta}_P=0$, $\alpha=0$) the score equals the entropy of the feature, i.e., the information gain of the question \citep{quinlan1986induction}; RPB thus generalizes the information-gain criterion with coverage, answerability and question difficulty. Third, it does not depend on the cluster size and never divides by zero. A higher RPB indicates a more useful question at the current turn.

\paragraph{Parameter estimation.}
$\kappa_P$, $p_i$, $m_P$ and $N_P$ are computed directly from the catalog and the current belief. Numerical features (e.g., price, memory and screen size) with more than 12 values are discretized into six quantile bins before scoring; otherwise each number is treated as a separate value and important features such as price are wrongly penalized. Text values are normalized (Persian and Latin digits, Arabic \emph{ye} and \emph{kaf}, zero-width non-joiners and white space). $\hat{\delta}_P$ is estimated from dialogues with Beta smoothing, $\hat{\delta}_P=(u_P+1)/(q_P+2)$, where $q_P$ is the number of times the feature was asked and $u_P$ the number of ``don't know'' or ``don't care'' answers. Without such data it is set by feature type: 0.05 for general features (at most 12 values), 0.5 for technical features (more values) and 0.95 for free-text descriptions (feature names containing ``descriptions'' or values longer than 40 characters on average), which a user cannot give as an answer. $\alpha$ is selected on validation clusters from $\{0, 0.1, 0.25, 0.5, 1\}$.

\paragraph{Static and dynamic use.}
With a uniform belief, $\kappa_P=n_c/n$, $p_i=x_i/n_c$ and $N_P=n_c$, where $n_c$ is the number of products that have the feature, and the features can be ranked once per cluster. In dynamic use, the score is recomputed on the updated belief after every answer, so features that overlap with earlier questions (e.g., the product series once the brand is known) automatically score lower. The belief is updated softly, so that a single wrong answer or extraction error does not eliminate the target product permanently:
\begin{equation}\label{eq:belief}
b_{t+1}(s) \propto b_t(s)\,L(a\mid s), \qquad
L(a\mid s) =
\begin{cases}
1-\epsilon & P(s)=a,\\
\epsilon/(m-1) & P(s)\neq a,\ s\notin S_P^{\varnothing},\\
\hat{p}_a & s\in S_P^{\varnothing},
\end{cases}
\end{equation}
where $a$ is the user's answer, $\epsilon$ the assumed answer or extraction error rate ($\epsilon=0.1$) and $\hat{p}_a$ the catalog frequency of value $a$; a ``don't know'' answer leaves the belief unchanged. Algorithm~\ref{alg:rpb} summarizes the procedure, which is used both for dataset construction (Section~\ref{subsec:planning}) and in the evaluation (Section~\ref{subsec:rpb-eval}).

\begin{algorithm}
\caption{Dynamic question selection with the RPB score}\label{alg:rpb}
\begin{algorithmic}[1]
\State $b_0(s) \propto \mathrm{popularity}(s)$; \ $\mathcal{Q} \gets \varnothing$
\For{$t = 0, 1, \dots, T-1$}
    \State $P^{\star} \gets \arg\max_{P \notin \mathcal{Q}} \mathrm{RPB}(P \mid b_t)$
    \State ask about $P^{\star}$ and receive the answer $a$; \ $\mathcal{Q} \gets \mathcal{Q} \cup \{P^{\star}\}$
    \If{$a$ is a valid value}
        \State update $b_{t+1}$ with Eq.~\eqref{eq:belief}
    \Else
        \State $b_{t+1} \gets b_t$
    \EndIf
    \State show the $k$ products with the highest $b_{t+1}$
\EndFor
\end{algorithmic}
\end{algorithm}

\subsection{Question planning}\label{subsec:planning}
For each target product, Algorithm~\ref{alg:rpb} is run on the target's cluster with the target's own values as the user's answers, $\alpha=0.1$ (selected in Section~\ref{subsec:rpb-eval}) and $\hat{\delta}_P$ set by feature type. If the target lacks the selected feature, the planned answer is ``don't know / doesn't matter'', which leaves the belief unchanged; such answers make the dialogues more realistic. Planning stops after five questions, or earlier once at least three questions have been asked and the target has the highest belief. The two products with the next-highest belief, preferably of the same product type as the target (the generic noun at the beginning of the title, e.g., ``pencil case''), become the alternatives that the assistant presents together with the target. The plan, i.e., the ordered features and the user's answers, is stored with each dialogue.

\subsection{Contextual augmentation}\label{subsec:augmentation}
To let the assistant compare products, each product is augmented with labels derived from its ten nearest neighbours within its cluster (cosine similarity of the MiniLM embeddings of title and description). The product is compared with this peer set on price, average rating and number of ratings, which yields its price position (budget, mid-range or premium) and labels such as ``good price'', ``premium price'', ``high rated'', ``popular choice'' or ``high value''.

\subsection{Dialogue generation}\label{subsec:generation}
We compared three strategies for turning the selected features into dialogues.

\paragraph{Single Prompt Strategy (SPS).} A single instruction asks the LLM to generate the entire conversation. The prompt describes the nature of the dialogue and the participants and contains the product information as a seed \citep{suresh-etal-2025-diasynth}.

\paragraph{Dual Prompt Strategy (DPS).} Two prompts, one per participant, define a role (seller or customer) and its goals, and two LLM instances converse \citep{duan-etal-2024-botchat}. This approach is more expensive, since the entire dialogue history has to be passed to each new prompt, which consumes many more tokens than SPS \citep{de-baer-etal-2025-single}.

\paragraph{Rewrite Template Strategy (RTS).} Dialogue templates that follow the feature-based filtering flow are filled with the selected features, and an LLM rewrites them into fluent, context-aware turns while preserving their intent and filtering logic \citep{shim2025tooldialmultiturndialoguegeneration}.

The strategy selected in Section~\ref{subsec:method-selection}, SPS with the RPB question plan and feature augmentation, was used for the released dataset. Its prompt contains the question plan, the target product (title, brand, price, rating, specifications, description, price position and comparison labels) and the two alternatives. It requires a \emph{requirement-driven} flow, which addresses the main error pattern found in the human assessment of the first version (Section~\ref{sec:human}): the user describes a need and names the generic product type, but not the brand, model or title of the product (for books, only ``a book''); the assistant asks about the planned features in order, one per turn; the user's answers follow the plan, with approximate, natural phrasing for numerical ranges; the assistant then presents the target and the alternatives with their exact prices and compares them with the comparison labels; the user asks one comparative question and finally chooses the target. The LLM must not invent any specification or price, and it tags every assistant turn that asks about a planned feature with the name of that feature.

\section{Experiments and results}\label{sec:experiments}

\subsection{Choosing the clustering configuration}\label{subsec:clustering}
\begin{figure}[ht]
    \centering
    \includegraphics[width=1\linewidth]{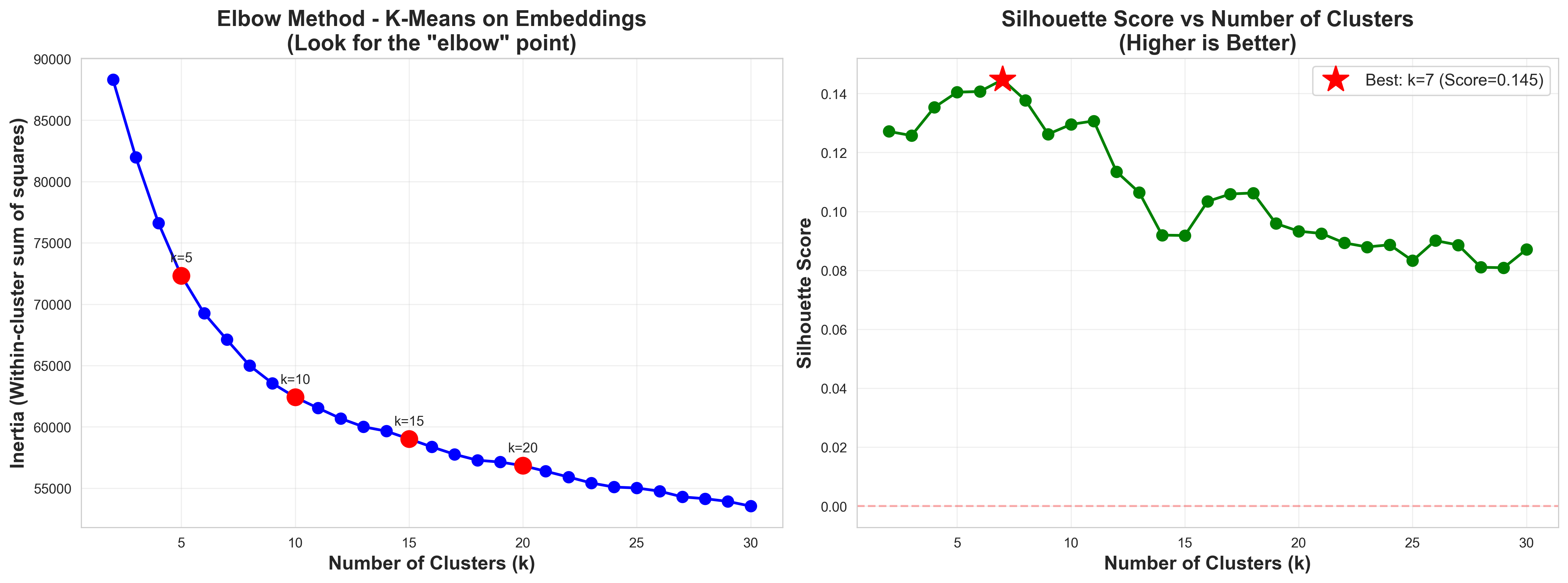}
    \caption{Elbow curve and Silhouette score of K-Means on MiniLM embeddings for 3 to 20 clusters.}
    \label{fig:lm-emb}
\end{figure}

\begin{figure}[ht]
    \centering
    \includegraphics[width=1\linewidth]{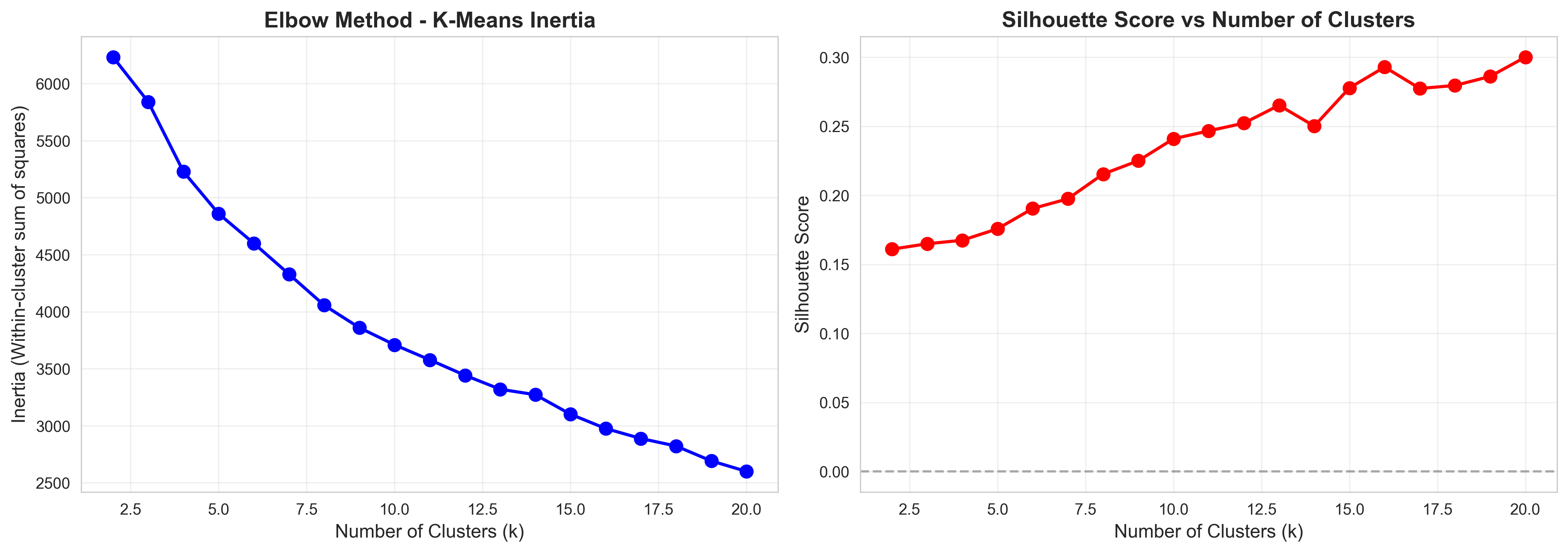}
    \caption{Elbow curve and Silhouette score of K-Means on TF-IDF vectors for 3 to 20 clusters.}
    \label{fig:tf-emb}
\end{figure}

We first verified that the approach recovers the seven original categories. Figures~\ref{fig:lm-emb} and~\ref{fig:tf-emb} show the elbow curves and Silhouette scores for 3 to 20 clusters with MiniLM and TF-IDF representations. With MiniLM embeddings, the Silhouette score clearly peaks at $k=7$ (0.145), matching the number of categories, whereas TF-IDF peaks at $k=20$; K-Means on MiniLM embeddings was therefore selected. Seven clusters, however, are far too coarse for comparing products, so the same configuration was applied with $k$ between 100 and 200, and $k=170$ was selected (Silhouette 0.105, average cluster size 76, minimum 9, maximum 270). All subsequent steps use these 170 clusters.


\subsection{Cluster homogeneity}
The RPB score compares the values of a feature within a cluster, so the products of a cluster should share their specification keys. We measured two quantities and compared them with a random partition into clusters of the same sizes: the \emph{key similarity}, the average Jaccard similarity between the specification-key sets of the products of a cluster, and the \emph{key coverage}, the average share of the cluster's products that have each of its keys. Table~\ref{tab:clustering-performance} shows that the K-Means clusters are far more homogeneous than random clusters, which is a prerequisite for computing the score within clusters.

\begin{table}[h]
\centering
\begin{tabular}{lrr}
\toprule
\textbf{Metric} & \textbf{Random} & \textbf{K-Means} \\
\midrule
Clusters & 170 & 170 \\
Average cluster size & 76.0 & 76.0 \\
Key similarity (mean) & 0.070 & 0.523 \\
Key similarity (std) & 0.007 & 0.217 \\
Key coverage (mean) & 0.061 & 0.363 \\
Key coverage (std) & 0.007 & 0.227 \\
\bottomrule
\end{tabular}
\caption{Homogeneity of the specification keys within K-Means and random clusters.}
\label{tab:clustering-performance}
\end{table}

\subsection{Evaluation of the RPB score on the Digikala catalog}
\label{subsec:rpb-eval}
In pre-written dialogues, the user's answers do not depend on the question the system chooses, so replaying them cannot show how much a question-selection criterion reduces the number of turns. We therefore evaluated the score of Eq.~\eqref{eq:rpb} on the real Digikala catalog with a simulated user: the catalog structure, i.e., the features, values, coverage and product popularity, is real, while the user's answers are simulated.

\paragraph{Data.} All 12,926 products were used with the 170 K-Means clusters. The features comprise all specification keys plus brand and price; price and other numerical features with more than 12 values were discretized into six quantile bins. Features for which more than 10\% of the products that have them carry several values at once (e.g., the list of smartwatch capabilities) were excluded, since they do not divide the products into disjoint groups; this removed 843 features, and 2,788 further features had a single value within their cluster. The remaining 5,029 features (29.6 per cluster on average) have low coverage: the median coverage is 0.39, and 55.5\% of the features are recorded for less than half of the products of their cluster. The number of ratings of each product plus one was used as its popularity.

\paragraph{User model.} The target product is drawn in proportion to popularity and hidden from the system. The user answers ``don't know'' with probability $\delta_P$; otherwise, the user gives the correct value with probability $1-\epsilon_P$, where $\epsilon_P=\min(0.4,\ 0.04\log_2 m)$, and on error names another value in proportion to its frequency. If the target lacks the feature in the catalog, the user names a value in proportion to the value frequencies. As the actual ``don't know'' rates for Digikala features are unavailable, two assumptions were examined: (1) a rate of 0.05 for features with at most 12 values and 0.5 for the others, and (2) a constant rate of 0.1 for all features.

\paragraph{Compared strategies.} Four question-selection strategies were compared: random feature selection; entropy with the default group counted as a value; entropy over the real values only; and the RPB score of Eq.~\eqref{eq:rpb}. Both entropy variants and the RPB score are recomputed on the updated belief after every answer. All strategies share the popularity prior, the belief update of Eq.~\eqref{eq:belief} with $\epsilon=0.1$, and common random numbers for the user's answers, so they differ only in the feature they select. $\hat{\delta}_P$ was estimated, without access to the true value, from ten pilot questions per feature with Beta smoothing.

\paragraph{Protocol and metrics.} Following common practice in conversational recommendation \citep{lei2020estimation,zhang2020evaluating}, the system shows its top $k$ products after every answer, and a dialogue succeeds the first time the target appears in this list. At most ten questions are allowed, and each feature is asked at most once. The success rate is the percentage of successful dialogues within this budget, and the average number of turns is the number of questions needed; failed dialogues count as 11 turns, and dialogues whose target is listed before the first question count as zero. The coefficient $\alpha$ of the RPB score was tuned on 42 randomly chosen clusters (25\%) by minimizing the average number of turns for $k=1$: $\alpha=0.1$ was selected under assumption 1 and $\alpha=0$ under assumption 2. The evaluation used the remaining 128 clusters with 15 target products each (1,920 dialogues per assumption), and 95\% confidence intervals were computed with a paired bootstrap with 2,000 resamples \citep{efron1993introduction}.

\begin{table}[h]
\centering
\resizebox{\textwidth}{!}{%
\begin{tabular}{llcccc}
\toprule
 & & \multicolumn{2}{c}{$k=1$} & \multicolumn{2}{c}{$k=3$} \\
\cmidrule(lr){3-4} \cmidrule(lr){5-6}
``Don't know'' rate & Strategy & Success (\%) & Avg.\ turns & Success (\%) & Avg.\ turns \\
\midrule
\multirow{4}{*}{Assumption 1}
 & Random & 72.7 & 5.14 & 87.3 & 3.04 \\
 & Entropy, default group & \textbf{90.1} & 2.93 & \textbf{96.2} & 1.76 \\
 & Entropy, real values & 83.9 & 3.86 & 93.9 & 2.20 \\
 & RPB & 89.6 & \textbf{2.83} & \textbf{96.2} & \textbf{1.52} \\
\midrule
\multirow{4}{*}{Assumption 2}
 & Random & 75.9 & 4.83 & 88.6 & 2.87 \\
 & Entropy, default group & \textbf{91.9} & \textbf{2.40} & \textbf{97.0} & 1.38 \\
 & Entropy, real values & 86.9 & 3.25 & 95.1 & 1.76 \\
 & RPB & 91.6 & 2.47 & 96.9 & \textbf{1.36} \\
\bottomrule
\end{tabular}%
}
\caption{Question-selection strategies on the real Digikala catalog with a simulated user (1,920 dialogues per assumption).}
\label{tab:rpb-real-results}
\end{table}

\begin{table}[h]
\centering
\small
\setlength{\tabcolsep}{4pt}
\begin{tabular}{lcccc}
\toprule
 & \multicolumn{2}{c}{$k=1$} & \multicolumn{2}{c}{$k=3$} \\
\cmidrule(lr){2-3} \cmidrule(lr){4-5}
Versus & Turns & Success (pp) & Turns & Success (pp) \\
\midrule
\multicolumn{5}{l}{\emph{Assumption 1}} \\[1pt]
\quad Entropy (default group) & \begin{tabular}[c]{@{}c@{}}$-0.11$\\[-1pt] {\scriptsize [$-0.19$, $-0.02$]}\end{tabular} & \begin{tabular}[c]{@{}c@{}}$-0.5$\\[-1pt] {\scriptsize [$-1.3$, $+0.2$]}\end{tabular} & \begin{tabular}[c]{@{}c@{}}$-0.24$\\[-1pt] {\scriptsize [$-0.30$, $-0.17$]}\end{tabular} & \begin{tabular}[c]{@{}c@{}}$+0.1$\\[-1pt] {\scriptsize [$-0.4$, $+0.5$]}\end{tabular} \\[3pt]
\addlinespace[2pt]
\multicolumn{5}{l}{\emph{Assumption 2}} \\[1pt]
\quad Entropy (default group) & \begin{tabular}[c]{@{}c@{}}$+0.07$\\[-1pt] {\scriptsize [$+0.02$, $+0.13$]}\end{tabular} & \begin{tabular}[c]{@{}c@{}}$-0.3$\\[-1pt] {\scriptsize [$-0.8$, $+0.2$]}\end{tabular} & \begin{tabular}[c]{@{}c@{}}$-0.02$\\[-1pt] {\scriptsize [$-0.06$, $+0.02$]}\end{tabular} & \begin{tabular}[c]{@{}c@{}}$-0.1$\\[-1pt] {\scriptsize [$-0.4$, $+0.3$]}\end{tabular} \\[3pt]
\bottomrule
\end{tabular}
\caption{Paired differences between the RPB score and the strongest baseline, entropy with the default group, on the real Digikala catalog, with 95\% confidence intervals below each estimate (negative turns and positive success favour RPB; pp = percentage points).}
\label{tab:rpb-real-diff}
\end{table}

The results are given in Tables~\ref{tab:rpb-real-results} and~\ref{tab:rpb-real-diff}. Under assumption 1, the RPB score needs significantly fewer turns than the strongest entropy variant (0.11 turns for $k=1$ and 0.24 for $k=3$), with a success rate that does not differ significantly.

Under assumption 2, where the ``don't know'' rate is the same for all features, the RPB score and entropy with the default group are practically equivalent: for $k=3$ the difference in turns is not significant, and for $k=1$ entropy needs 0.07 fewer turns, a small but significant difference, with no significant difference in success. Most of the advantage over entropy therefore comes from the answerability term and vanishes when that rate is constant. Entropy over the real values, which ignores coverage, is significantly worse under both assumptions (1.04 more turns for $k=1$ under assumption 1).

An ablation under assumption 1 confirms the role of coverage: removing $\kappa_P$ lowers the success rate for $k=1$ from 89.6\% to 82.7\% and raises the average number of turns from 2.83 to 3.91, whereas removing $(1-\hat{\delta}_P)$ has a smaller effect (89.5\% and 2.89 turns). The cardinality penalty plays a minor role: with $\alpha=0$, the results are almost unchanged (90.1\% and 2.81 turns).

Overall, on the real Digikala catalog the RPB score performs on par with the strongest entropy variant and is better when the difficulty of answering differs between features.

\subsection{Selection of the generation method}\label{subsec:method-selection}
To select the generation strategy, five settings were compared, crossing the three strategies of Section~\ref{subsec:generation} with two ways of choosing the features: the question plan of the RPB score combined with the contextual augmentation (RPB + FA), and five randomly chosen features without augmentation (Random), each target with the same two alternatives. DPS was run with RPB + FA only. Each setting was run with three LLMs of the free Gemini API, Gemini~3.1 Flash-Lite and Gemini~3.5 Flash-Lite \citep{gemini2023} and the open-weight Gemma~4 26B-A4B, a successor of Gemma~3 \citep{gemma2025}, on the same ten products, at least one from every category and none of them in the released dataset. Of the 150 dialogues, 148 were obtained; two Gemma dialogues (SPS + Random) failed because the server was repeatedly overloaded. Each dialogue was scored from 1 to 10 by an LLM judge, Gemini~3.1 Flash-Lite (preview), on the six criteria of Table~\ref{tab:dialogue-metrics-description}.

Table~\ref{tab:model-comparison} shows the average scores of the three LLMs over all settings. Gemini~3.1 Flash-Lite obtained the highest scores on all six criteria, followed closely by Gemma~4; Gemini~3.5 Flash-Lite was lower, mainly in realism (6.98 vs.\ 8.08). Since the judge belongs to the same model line as Gemini~3.1 Flash-Lite, part of this lead may come from self-preference \citep{panickssery2024llm}; the ranking of Gemma~4 above Gemini~3.5 Flash-Lite is not affected by this bias.

Table~\ref{tab:model-kind-comparison} compares the settings. All settings obtained similar overall scores (7.90--8.90), and no strategy was best with every LLM. The RPB plans needed 3.3 questions on average, against five for the random plans, so their dialogues were shorter. Pooled over the LLMs and paired by product, RPB + FA minus Random gave an overall difference of $-0.25$ (95\% bootstrap CI $[-0.68, 0.18]$) for SPS and $+0.10$ ($[-0.33, 0.53]$) for RTS, neither significant, while coverage was lower with RPB + FA ($-0.79$, $[-1.18, -0.43]$ for SPS; $-0.43$, $[-0.77, -0.07]$ for RTS). The judge thus rewards dialogues that mention more features, whereas the RPB plans aim to identify the target with as few questions as possible, the property measured in Section~\ref{subsec:rpb-eval}; at equal overall quality, a shorter dialogue is preferable for the user. Among the strategies, SPS needs a single LLM call per dialogue and no hand-written templates, whereas DPS needed 8.7 calls on average (one per message) without a consistent quality gain. SPS with RPB + FA was therefore used for the released dataset. Because the comparison uses ten products and a single judge, it shows that the settings give similar quality rather than precise differences between them.

\begin{table}[h]
\centering
\begin{tabular}{lccc}
\toprule
Criterion & Gemini~3.1 FL & Gemini~3.5 FL & Gemma~4 \\
\midrule
Fluency & \textbf{9.80} & 9.42 & 9.75 \\
Accuracy & \textbf{8.82} & 8.42 & 8.73 \\
Realism & \textbf{8.08} & 6.98 & 7.79 \\
Coverage & \textbf{8.58} & 8.24 & 8.50 \\
Helpfulness & \textbf{8.84} & 8.24 & 8.75 \\
Overall & \textbf{8.70} & 8.16 & 8.56 \\
\bottomrule
\end{tabular}
\caption{Average LLM-judge scores (1--10) of the three LLMs over all five generation settings (ten products; 50, 50 and 48 dialogues). FL: Flash-Lite; Gemma~4: Gemma~4 26B-A4B.}
\label{tab:model-comparison}
\end{table}

\begin{table*}[h]
\centering
\resizebox{\textwidth}{!}{%
\begin{tabular}{lccccccccccc}
\toprule
Setting & Questions & \multicolumn{3}{c}{Overall} & \multicolumn{3}{c}{Accuracy} & \multicolumn{3}{c}{Coverage} \\
\cmidrule(lr){3-5} \cmidrule(lr){6-8} \cmidrule(lr){9-11}
 & & 3.1 FL & 3.5 FL & Gemma~4 & 3.1 FL & 3.5 FL & Gemma~4 & 3.1 FL & 3.5 FL & Gemma~4 \\
\midrule
SPS + RPB + FA & 3.3 & 8.50 & 8.20 & 8.40 & 8.50 & 8.50 & 8.20 & 8.30 & 8.00 & 8.20 \\
SPS + Random & 5 & \textbf{8.90} & \textbf{8.50} & 8.50 & \textbf{9.00} & \textbf{8.70} & 8.75 & \textbf{9.10} & \textbf{8.70} & \textbf{9.00} \\
RTS + RPB + FA & 3.3 & 8.60 & 8.30 & 8.60 & \textbf{9.00} & 8.50 & 8.90 & 8.40 & 8.20 & 8.30 \\
RTS + Random & 5 & 8.60 & 7.90 & \textbf{8.70} & 8.80 & 8.20 & \textbf{9.10} & 8.80 & 8.50 & 8.90 \\
DPS + RPB + FA & 3.3 & \textbf{8.90} & 7.90 & 8.60 & 8.80 & 8.20 & 8.70 & 8.30 & 7.80 & 8.20 \\
\bottomrule
\end{tabular}}
\caption{LLM-judge scores of the generation settings with three LLMs (ten products per cell; eight for Gemma~4 with SPS + Random). Questions: average number of planned questions. RTS: Rewrite Template Strategy; SPS: Single Prompt Strategy; DPS: Dual Prompt Strategy; RPB + FA: question plan of the RPB score plus contextual augmentation; Random: five randomly selected features without augmentation; FL: Flash-Lite.}
\label{tab:model-kind-comparison}
\end{table*}

\begin{table}[h]
\small
\centering
\begin{tabular}{p{1.8cm}p{5.2cm}}
\toprule
\textbf{Criterion} & \textbf{Description and examples} \\
\midrule
\textbf{Fluency} & Naturalness, smoothness, grammatical correctness. \\
& \textit{High:} sentences flow well, no awkward phrasing. \\
& \textit{Low:} broken grammar, repetitive wording. \\
\addlinespace
\textbf{Accuracy} & Consistency with the product information. \\
& \textit{High:} features and specifications match the product. \\
& \textit{Low:} wrong specifications or price, hallucinated features. \\
\addlinespace
\textbf{Realism} & Believability as a real conversation. \\
& \textit{High:} realistic questions and clarifications. \\
& \textit{Low:} robotic tone, artificial behaviour. \\
\addlinespace
\textbf{Coverage} & Coverage of the important product aspects. \\
& \textit{High:} most important features discussed. \\
& \textit{Low:} critical features missed. \\
\addlinespace
\textbf{Helpfulness} & Effectiveness for the purchase decision. \\
& \textit{High:} clarifies requirements, compares options. \\
& \textit{Low:} vague answers, leaves the user uncertain. \\
\addlinespace
\textbf{Overall} & Global quality combining all criteria. \\
& \textit{High:} strong fluency, accurate content. \\
& \textit{Low:} significant problems in language or facts. \\
\bottomrule
\end{tabular}
\caption{Criteria of the LLM-based dialogue evaluation.}
\label{tab:dialogue-metrics-description}
\end{table}

\section{Human assessment of the first version}\label{sec:human}
The first version of the dataset was built before the question plans of Section~\ref{subsec:planning} were introduced: it consists of 26 dialogues generated with GPT-4o \citep{openai2024gpt4o} and SPS with feature augmentation, using the highest-scoring features of each cluster, ranked once per cluster. These dialogues were reviewed qualitatively by the authors, who read all 26 dialogues and compared every statement about a product with its catalog entry. Besides confirming the high fluency found by the LLM judge, the review revealed one systematic error pattern: an inconsistency between two conversational flows.

\begin{itemize}
    \item \emph{Product-specific inquiry}: the user begins by naming a specific product and asks for its details.
    \item \emph{Requirement-driven search}: the user begins by describing needs or desired attributes, and the assistant searches for matching products.
\end{itemize}
The dataset is meant to follow the second, search-oriented flow, but several dialogues followed the first. Dialogues intended to be exploratory then fixed prematurely on one product: the questions concerned the features of that product instead of the user's requirements, and the conclusion invariably favoured it, leaving the suggested alternatives unselected. The released version addresses this pattern with the question plan and an explicitly requirement-driven prompt (Section~\ref{subsec:generation}); Section~\ref{sec:dataset} measures how often the first user message still names the target.

\section{The released dataset}\label{sec:dataset}
\begin{figure}[t]
\centering
\includegraphics[width=\textwidth]{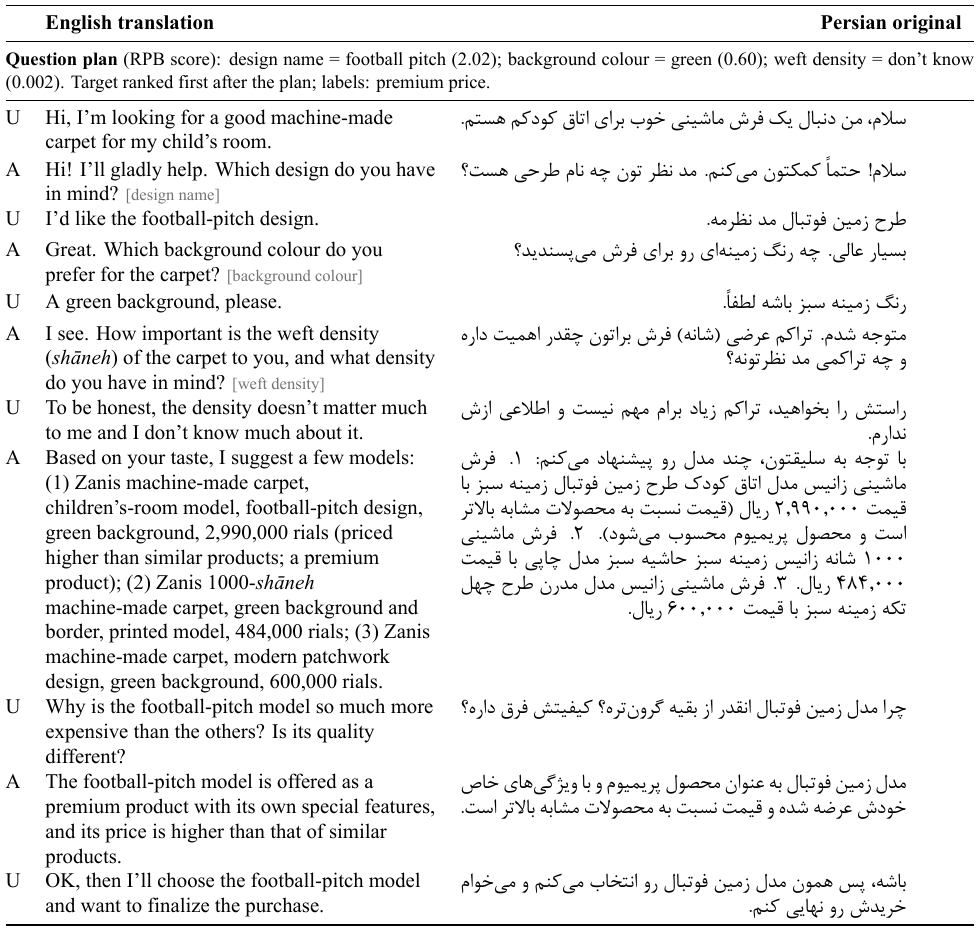}
\caption{An example dialogue of the released dataset (\texttt{pcps-7944}, generated by Gemini~3.5 Flash-Lite) with its English translation. U: user; A: assistant; grey brackets: the feature tag of each planned question. The third planned feature is missing for the target, so the user answers ``don't know''; the assistant then presents the target and the two alternatives with their exact prices and the comparison label.}
\label{fig:example}
\end{figure}

\begin{sloppypar}The released version of the dataset was generated with the RPB score in dynamic form (Section~\ref{subsec:planning}) and the free tier of the Google Gemini API \citep{gemini2023} (\texttt{gemini-3.1-flash-lite}: 491, \texttt{gemini-3.5-flash-lite}: 450, \texttt{gemini-3-flash-preview}: 37, \texttt{gemini-3.5-flash}: 22), so that the dataset can be reproduced and extended without cost. The targets were sampled from all 170 clusters, in proportion to the cluster sizes and uniformly within each cluster, so that the dataset also covers products with few ratings. Each record contains the dialogue, the question plan with the user's answers, the feature tag of every assistant question, the target product and the two alternatives (with their Digikala identifiers), the comparison labels and the generating model. Table~\ref{tab:dataset-stats} summarizes the dataset. Figure~\ref{fig:example} shows one dialogue with its question plan and feature tags.\end{sloppypar}

\begin{table}[h]
\centering
\begin{tabular}{lr}
\toprule
Dialogues & 1,000 \\
Clusters / categories covered & 170 / 7 \\
Messages (total) & 12,540 \\
Messages per dialogue (mean) & 12.5 \\
Words per message (mean) & 16.3 \\
Planned questions per dialogue (mean) & 3.75 \\
``Don't know'' answers among planned questions & 17.6\% \\
Distinct features asked & 186 \\
\bottomrule
\end{tabular}
\caption{Statistics of the released dataset.}
\label{tab:dataset-stats}
\end{table}

\paragraph{Question efficiency.} With the true answers of the target, the RPB score placed the target first among the products of its cluster in 73.6\% of the plans after 3.75 questions on average.

\paragraph{Automatic checks.} 89.2\% of the planned categorical answers appear verbatim in the dialogue (numerical ranges are paraphrased by design and were not checked), 99.97\% of the planned features are tagged on an assistant turn, and in 88.7\% of the dialogues the first user message does not name the target product (fewer than half of its distinctive title words), i.e., the dialogue is requirement-driven.

\paragraph{LLM judge.} A random sample of 100 dialogues of the released version and the 26 dialogues of the first version were scored with the criteria of Table~\ref{tab:dialogue-metrics-description} by Gemini~3.1 Flash-Lite as judge (Table~\ref{tab:judge}; 119 of the 125 scores, the others by Gemini~3.5 Flash-Lite and Gemini~3.5 Flash after the daily quota was exhausted; one judge output of the sample could not be parsed, leaving 99). The released version obtained a higher accuracy score (7.87 vs.\ 6.85), in line with the grounding of every answer in the question plan, but a lower realism score (7.18 vs.\ 8.27), which we attribute to the smaller free models and to dialogues that follow a fixed plan; the other criteria were similar. As the main judge also generated 491 of the released dialogues and belongs to the same model family as the other generators, its scores may favour that version \citep{panickssery2024llm}; they serve as a consistency check rather than as a comparison of the two versions.

\begin{table}[h]
\centering
\begin{tabular}{lcc}
\toprule
Criterion & First version & Released version \\
 & (GPT-4o) & (Gemini) \\
\midrule
Fluency & 9.81 & 9.49 \\
Accuracy & 6.85 & 7.87 \\
Realism & 8.27 & 7.18 \\
Coverage & 8.58 & 8.21 \\
Helpfulness & 8.12 & 8.11 \\
Overall & 7.92 & 8.08 \\
\bottomrule
\end{tabular}
\caption{LLM-judge scores (1--10) of the two dataset versions.}
\label{tab:judge}
\end{table}

\paragraph{Comparison with existing datasets.} Table~\ref{tab:dataset-comparison} compares the released dataset with e-commerce dialogue datasets that are built around products. It is the only Persian dataset grounded in a large real catalog, and the only one that stores the question plan of every dialogue together with the feature asked at each turn. PerSHOP \citep{mahmoudi2024pershoppersiandataset} contains human-written dialogues with intent and slot annotations, but over a catalog of 750 products; Wizard of Shopping \citep{li2025wizardshoppingtargetorientedecommerce} is larger and also generated from a plan over a real catalog, but it is in English and has not been released. Our dataset and Wizard of Shopping are the only ones that can be generated automatically, so they can be extended to new products or categories at little cost; ours uses only free models, so this costs nothing.

\begin{table}[h]
\centering
\small
\resizebox{\textwidth}{!}{%
\begin{tabular}{lllrrlcc}
\toprule
Dataset & Language & Construction & Dialogues & Categories & Catalog & Automatic & Public \\
\midrule
MG-ShopDial \citep{bernard2023mgshopdial} & English & crowd-sourced & 64 & -- & curated & no & yes \\
Wizard of Shopping \citep{li2025wizardshoppingtargetorientedecommerce} & English & GPT-4, decision-tree plan & 3,600 & 3 & 236k (Amazon) & yes & no \\
PerSHOP \citep{mahmoudi2024pershoppersiandataset} & Persian & crowd-sourced, human--human & 1,061 & 14 & 750 & no & yes \\
Ours & Persian & free Gemini LLMs, RPB plan & 1,000 & 7 (170 clusters) & 12,926 (Digikala) & yes & yes \\
\bottomrule
\end{tabular}}
\caption{Comparison with product-grounded e-commerce dialogue datasets. Catalog: number of products from which the dialogues are drawn. Automatic: new dialogues can be generated from a catalog without human writers or annotators. Turn-level annotations are intents and 36 slot types for PerSHOP and the planned feature of every assistant question for our dataset.}
\label{tab:dataset-comparison}
\end{table}

\section{Limitations}\label{sec:limitations}
The LLM judge was used both to select the generation method and to check the released dataset, and the method preferred by an LLM may differ from the one preferred by people, and the judge of the method comparison belongs to the same model line as one of the compared LLMs; the human assessment was a qualitative review of the first version by the authors, without independent annotators or agreement scores. The RPB score was evaluated with a simulated user whose ``don't know'' and error rates were assumed, and its advantage over entropy is small and depends on how much the difficulty of answering differs between features. The question plans use the target's catalog values as the user's answers, so the dialogues contain no answer errors, and the ``don't know'' rates by feature type are heuristic. Multi-valued features were excluded. Finally, the dataset covers seven categories of one marketplace.

\section{Conclusion}\label{sec:conclusion}
This study introduced a framework for constructing Persian conversational product search datasets grounded in a real e-commerce catalog, addressing the data scarcity that has limited Persian dialogue systems. The framework clusters 12,926 Digikala products into 170 homogeneous clusters, plans the questions of every dialogue with the RPB score, augments products with comparisons to their neighbours and verbalizes the plan with an LLM.

The RPB score accounts for feature coverage, answerability and the number of values, and is recomputed after every answer. On the full Digikala catalog with a simulated user, the RPB score required 0.11 fewer turns than the strongest entropy baseline when the difficulty of answering differs between features, and was practically equivalent to it otherwise.

Single-prompt, dual-prompt and template-rewrite generation obtained similar LLM-judge scores with three free LLMs, and RPB plans gave dialogues of equal overall quality with fewer questions than random features; the single-prompt strategy was chosen because it needs one LLM call per dialogue. A human assessment of the first version revealed a drift from requirement-driven to product-specific flows, which the released version addresses with explicit question plans. The released dataset of 1,000 dialogues was generated with a free LLM API and covers all 170 clusters.

Future work includes estimating the ``don't know'' rates from real dialogues, evaluating the score with real users, a quantitative human evaluation of the released version with independent annotators, simulating answer errors and multi-valued features in the plans, and extending the dataset to further categories.

\section*{Data and code availability}
The product catalog (public product pages of Digikala), the 170 clusters, the question plans and the dialogue dataset are available at \url{https://github.com/farinfr/persian-cps-dialogues}. The dialogue dataset is released under the Creative Commons Attribution-NonCommercial 4.0 International license (CC BY-NC 4.0); product data remain subject to the terms of the Digikala platform and are shared for research purposes only. The code for crawling, clustering, the RPB score, the simulation and the dataset generation is available in the same repository under the MIT license.

\begin{acks}
No funding was received for conducting this study. The authors have no competing interests to declare.
\end{acks}

\bibliographystyle{ACM-Reference-Format}
\bibliography{references}


\begin{thebibliography}{66}


\ifx \showCODEN    \undefined \def \showCODEN     #1{\unskip}     \fi
\ifx \showISBNx    \undefined \def \showISBNx     #1{\unskip}     \fi
\ifx \showISBNxiii \undefined \def \showISBNxiii  #1{\unskip}     \fi
\ifx \showISSN     \undefined \def \showISSN      #1{\unskip}     \fi
\ifx \showLCCN     \undefined \def \showLCCN      #1{\unskip}     \fi
\ifx \shownote     \undefined \def \shownote      #1{#1}          \fi
\ifx \showarticletitle \undefined \def \showarticletitle #1{#1}   \fi
\ifx \showURL      \undefined \def \showURL       {\relax}        \fi
\providecommand\bibfield[2]{#2}
\providecommand\bibinfo[2]{#2}
\providecommand\natexlab[1]{#1}
\providecommand\showeprint[2][]{arXiv:#2}

\bibitem[Abaskohi et~al\mbox{.}(2024)]%
        {abaskohi2024benchmarking}
\bibfield{author}{\bibinfo{person}{Amirhossein Abaskohi}, \bibinfo{person}{Sara
  Baruni}, \bibinfo{person}{Mostafa Masoudi}, \bibinfo{person}{Nesa Abbasi},
  \bibinfo{person}{Mohammad~Hadi Babalou}, \bibinfo{person}{Ali Edalat},
  \bibinfo{person}{Sepehr Kamahi}, \bibinfo{person}{Samin Mahdizadeh~Sani},
  \bibinfo{person}{Nikoo Naghavian}, \bibinfo{person}{Danial Namazifard},
  \bibinfo{person}{Pouya Sadeghi}, {and} \bibinfo{person}{Yadollah
  Yaghoobzadeh}.} \bibinfo{year}{2024}\natexlab{}.
\newblock \showarticletitle{Benchmarking Large Language Models for {P}ersian: A
  Preliminary Study Focusing on {C}hat{GPT}}. In
  \bibinfo{booktitle}{\emph{Proceedings of the 2024 Joint International
  Conference on Computational Linguistics, Language Resources and Evaluation
  (LREC-COLING 2024)}}, \bibfield{editor}{\bibinfo{person}{Nicoletta
  Calzolari}, \bibinfo{person}{Min-Yen Kan}, \bibinfo{person}{Veronique Hoste},
  \bibinfo{person}{Alessandro Lenci}, \bibinfo{person}{Sakriani Sakti}, {and}
  \bibinfo{person}{Nianwen Xue}} (Eds.). \bibinfo{publisher}{ELRA and ICCL},
  \bibinfo{address}{Torino, Italia}, \bibinfo{pages}{2189--2203}.
\newblock
\urldef\tempurl%
\url{https://aclanthology.org/2024.lrec-main.197/}
\showURL{%
\tempurl}


\bibitem[Abbasi et~al\mbox{.}(2025)]%
        {abbasi2025hamraz}
\bibfield{author}{\bibinfo{person}{Mohammad~Amin Abbasi},
  \bibinfo{person}{Farnaz~Sadat Mirnezami}, \bibinfo{person}{Ali Neshati},
  {and} \bibinfo{person}{Hassan Naderi}.} \bibinfo{year}{2025}\natexlab{}.
\newblock \bibinfo{title}{{HamRaz}: A Culture-Based {P}ersian Conversation
  Dataset for Person-Centered Therapy Using {LLM} Agents}.
\newblock
\showeprint[arxiv]{2502.05982}
\urldef\tempurl%
\url{https://arxiv.org/abs/2502.05982}
\showURL{%
\tempurl}


\bibitem[Aliannejadi et~al\mbox{.}(2021)]%
        {aliannejadi2021building}
\bibfield{author}{\bibinfo{person}{Mohammad Aliannejadi},
  \bibinfo{person}{Julia Kiseleva}, \bibinfo{person}{Aleksandr Chuklin},
  \bibinfo{person}{Jeff Dalton}, {and} \bibinfo{person}{Mikhail Burtsev}.}
  \bibinfo{year}{2021}\natexlab{}.
\newblock \showarticletitle{Building and Evaluating Open-Domain Dialogue
  Corpora with Clarifying Questions}. In \bibinfo{booktitle}{\emph{Proceedings
  of the 2021 Conference on Empirical Methods in Natural Language Processing
  (EMNLP)}}. \bibinfo{pages}{4473--4484}.
\newblock
\href{https://doi.org/10.18653/v1/2021.emnlp-main.367}{doi:\nolinkurl{10.18653/v1/2021.emnlp-main.367}}


\bibitem[Aliannejadi et~al\mbox{.}(2019)]%
        {aliannejadi2019asking}
\bibfield{author}{\bibinfo{person}{Mohammad Aliannejadi},
  \bibinfo{person}{Hamed Zamani}, \bibinfo{person}{Fabio Crestani}, {and}
  \bibinfo{person}{W.~Bruce Croft}.} \bibinfo{year}{2019}\natexlab{}.
\newblock \showarticletitle{Asking Clarifying Questions in Open-Domain
  Information-Seeking Conversations}. In \bibinfo{booktitle}{\emph{Proceedings
  of the 42nd International ACM SIGIR Conference on Research and Development in
  Information Retrieval}}. \bibinfo{pages}{475--484}.
\newblock
\href{https://doi.org/10.1145/3331184.3331265}{doi:\nolinkurl{10.1145/3331184.3331265}}


\bibitem[Bernard and Balog(2023)]%
        {bernard2023mgshopdial}
\bibfield{author}{\bibinfo{person}{Nolwenn Bernard} {and}
  \bibinfo{person}{Krisztian Balog}.} \bibinfo{year}{2023}\natexlab{}.
\newblock \showarticletitle{{MG-ShopDial}: A Multi-Goal Conversational Dataset
  for e-Commerce}. In \bibinfo{booktitle}{\emph{Proceedings of the 46th
  International ACM SIGIR Conference on Research and Development in Information
  Retrieval}} \emph{(\bibinfo{series}{SIGIR '23})}. \bibinfo{publisher}{ACM},
  \bibinfo{pages}{2775--2785}.
\newblock
\href{https://doi.org/10.1145/3539618.3591883}{doi:\nolinkurl{10.1145/3539618.3591883}}


\bibitem[Bi et~al\mbox{.}(2021)]%
        {bi2021asking}
\bibfield{author}{\bibinfo{person}{Keping Bi}, \bibinfo{person}{Qingyao Ai},
  {and} \bibinfo{person}{W.~Bruce Croft}.} \bibinfo{year}{2021}\natexlab{}.
\newblock \showarticletitle{Asking Clarifying Questions Based on Negative
  Feedback in Conversational Search}. In \bibinfo{booktitle}{\emph{Proceedings
  of the 2021 ACM SIGIR International Conference on Theory of Information
  Retrieval}} \emph{(\bibinfo{series}{ICTIR '21})}. \bibinfo{publisher}{ACM},
  \bibinfo{pages}{157--166}.
\newblock
\href{https://doi.org/10.1145/3471158.3472232}{doi:\nolinkurl{10.1145/3471158.3472232}}


\bibitem[Bi et~al\mbox{.}(2019)]%
        {bi2019conversational}
\bibfield{author}{\bibinfo{person}{Keping Bi}, \bibinfo{person}{Qingyao Ai},
  \bibinfo{person}{Yongfeng Zhang}, {and} \bibinfo{person}{W.~Bruce Croft}.}
  \bibinfo{year}{2019}\natexlab{}.
\newblock \showarticletitle{Conversational Product Search Based on Negative
  Feedback}. In \bibinfo{booktitle}{\emph{Proceedings of the 28th ACM
  International Conference on Information and Knowledge Management (CIKM)}}.
  \bibinfo{pages}{359--368}.
\newblock
\href{https://doi.org/10.1145/3357384.3357939}{doi:\nolinkurl{10.1145/3357384.3357939}}


\bibitem[Borhanifard et~al\mbox{.}(2020)]%
        {9345639}
\bibfield{author}{\bibinfo{person}{Zeinab Borhanifard},
  \bibinfo{person}{Hossein Basafa}, \bibinfo{person}{Seyedeh~Zahra Razavi},
  {and} \bibinfo{person}{Heshaam Faili}.} \bibinfo{year}{2020}\natexlab{}.
\newblock \showarticletitle{Persian Language Understanding in Task-Oriented
  Dialogue System for Online Shopping}. In \bibinfo{booktitle}{\emph{2020 11th
  International Conference on Information and Knowledge Technology (IKT)}}.
  \bibinfo{pages}{79--84}.
\newblock
\href{https://doi.org/10.1109/IKT51791.2020.9345639}{doi:\nolinkurl{10.1109/IKT51791.2020.9345639}}


\bibitem[Calinski and Harabasz(1974)]%
        {calinski1974dendrite}
\bibfield{author}{\bibinfo{person}{Tadeusz Calinski} {and}
  \bibinfo{person}{Jerzy Harabasz}.} \bibinfo{year}{1974}\natexlab{}.
\newblock \showarticletitle{A dendrite method for cluster analysis}.
\newblock \bibinfo{journal}{\emph{Communications in Statistics-theory and
  Methods}} \bibinfo{volume}{3}, \bibinfo{number}{1} (\bibinfo{year}{1974}),
  \bibinfo{pages}{1--27}.
\newblock


\bibitem[Chen et~al\mbox{.}(2023)]%
        {chen2023places}
\bibfield{author}{\bibinfo{person}{Maximillian Chen},
  \bibinfo{person}{Alexandros Papangelis}, \bibinfo{person}{Chenyang Tao},
  \bibinfo{person}{Seokhwan Kim}, \bibinfo{person}{Andy Rosenbaum},
  \bibinfo{person}{Yang Liu}, \bibinfo{person}{Zhou Yu}, {and}
  \bibinfo{person}{Dilek Hakkani-Tur}.} \bibinfo{year}{2023}\natexlab{}.
\newblock \showarticletitle{{PLACES}: Prompting Language Models for Social
  Conversation Synthesis}. In \bibinfo{booktitle}{\emph{Findings of the
  Association for Computational Linguistics: EACL 2023}},
  \bibfield{editor}{\bibinfo{person}{Andreas Vlachos} {and}
  \bibinfo{person}{Isabelle Augenstein}} (Eds.).
  \bibinfo{publisher}{Association for Computational Linguistics},
  \bibinfo{address}{Dubrovnik, Croatia}, \bibinfo{pages}{844--868}.
\newblock
\href{https://doi.org/10.18653/v1/2023.findings-eacl.63}{doi:\nolinkurl{10.18653/v1/2023.findings-eacl.63}}


\bibitem[Christakopoulou et~al\mbox{.}(2016)]%
        {christakopoulou2016towards}
\bibfield{author}{\bibinfo{person}{Konstantina Christakopoulou},
  \bibinfo{person}{Filip Radlinski}, {and} \bibinfo{person}{Katja Hofmann}.}
  \bibinfo{year}{2016}\natexlab{}.
\newblock \showarticletitle{Towards Conversational Recommender Systems}. In
  \bibinfo{booktitle}{\emph{Proceedings of the 22nd ACM SIGKDD International
  Conference on Knowledge Discovery and Data Mining}}.
  \bibinfo{pages}{815--824}.
\newblock
\href{https://doi.org/10.1145/2939672.2939746}{doi:\nolinkurl{10.1145/2939672.2939746}}


\bibitem[Davies and Bouldin(1979)]%
        {davies1979cluster}
\bibfield{author}{\bibinfo{person}{David~L Davies} {and}
  \bibinfo{person}{Donald~W Bouldin}.} \bibinfo{year}{1979}\natexlab{}.
\newblock \showarticletitle{A cluster separation measure}.
\newblock \bibinfo{journal}{\emph{IEEE transactions on pattern analysis and
  machine intelligence}} \bibinfo{number}{2} (\bibinfo{year}{1979}),
  \bibinfo{pages}{224--227}.
\newblock


\bibitem[De~Baer et~al\mbox{.}(2025)]%
        {de-baer-etal-2025-single}
\bibfield{author}{\bibinfo{person}{Joachim De~Baer}, \bibinfo{person}{A.~Seza
  Do{\u{g}}ru{\"o}z}, \bibinfo{person}{Thomas Demeester}, {and}
  \bibinfo{person}{Chris Develder}.} \bibinfo{year}{2025}\natexlab{}.
\newblock \showarticletitle{Single- vs. Dual-Prompt Dialogue Generation with
  {LLM}s for Job Interviews in Human Resources}. In
  \bibinfo{booktitle}{\emph{Proceedings of the Fourth Workshop on Generation,
  Evaluation and Metrics (GEM{\texttwosuperior})}},
  \bibfield{editor}{\bibinfo{person}{Ofir Arviv}, \bibinfo{person}{Miruna
  Clinciu}, \bibinfo{person}{Kaustubh Dhole}, \bibinfo{person}{Rotem Dror},
  \bibinfo{person}{Sebastian Gehrmann}, \bibinfo{person}{Eliya Habba},
  \bibinfo{person}{Itay Itzhak}, \bibinfo{person}{Simon Mille},
  \bibinfo{person}{Yotam Perlitz}, \bibinfo{person}{Enrico Santus},
  \bibinfo{person}{Jo{\~a}o Sedoc}, \bibinfo{person}{Michal Shmueli~Scheuer},
  \bibinfo{person}{Gabriel Stanovsky}, {and} \bibinfo{person}{Oyvind Tafjord}}
  (Eds.). \bibinfo{publisher}{Association for Computational Linguistics},
  \bibinfo{address}{Vienna, Austria and virtual meeting},
  \bibinfo{pages}{947--957}.
\newblock
\showISBNx{979-8-89176-261-9}
\urldef\tempurl%
\url{https://aclanthology.org/2025.gem-1.74/}
\showURL{%
\tempurl}


\bibitem[{Digikala}(2025)]%
        {digikala}
\bibfield{author}{\bibinfo{person}{{Digikala}}.}
  \bibinfo{year}{2025}\natexlab{}.
\newblock \bibinfo{title}{Digikala Online Marketplace}.
\newblock \bibinfo{howpublished}{\url{https://www.digikala.com}}.
\newblock
\newblock
\shownote{Accessed: 2025-10-12}.


\bibitem[Duan et~al\mbox{.}(2024)]%
        {duan-etal-2024-botchat}
\bibfield{author}{\bibinfo{person}{Haodong Duan}, \bibinfo{person}{Jueqi Wei},
  \bibinfo{person}{Chonghua Wang}, \bibinfo{person}{Hongwei Liu},
  \bibinfo{person}{Yixiao Fang}, \bibinfo{person}{Songyang Zhang},
  \bibinfo{person}{Dahua Lin}, {and} \bibinfo{person}{Kai Chen}.}
  \bibinfo{year}{2024}\natexlab{}.
\newblock \showarticletitle{{B}ot{C}hat: Evaluating {LLM}s' Capabilities of
  Having Multi-Turn Dialogues}. In \bibinfo{booktitle}{\emph{Findings of the
  Association for Computational Linguistics: NAACL 2024}},
  \bibfield{editor}{\bibinfo{person}{Kevin Duh}, \bibinfo{person}{Helena
  Gomez}, {and} \bibinfo{person}{Steven Bethard}} (Eds.).
  \bibinfo{publisher}{Association for Computational Linguistics},
  \bibinfo{address}{Mexico City, Mexico}, \bibinfo{pages}{3184--3200}.
\newblock
\href{https://doi.org/10.18653/v1/2024.findings-naacl.201}{doi:\nolinkurl{10.18653/v1/2024.findings-naacl.201}}


\bibitem[Efron and Tibshirani(1993)]%
        {efron1993introduction}
\bibfield{author}{\bibinfo{person}{Bradley Efron} {and}
  \bibinfo{person}{Robert~J. Tibshirani}.} \bibinfo{year}{1993}\natexlab{}.
\newblock \bibinfo{booktitle}{\emph{An Introduction to the Bootstrap}}.
\newblock \bibinfo{publisher}{Chapman \& Hall}, \bibinfo{address}{New York}.
\newblock


\bibitem[Ester et~al\mbox{.}(1996)]%
        {ester1996density}
\bibfield{author}{\bibinfo{person}{Martin Ester}, \bibinfo{person}{Hans-Peter
  Kriegel}, \bibinfo{person}{J{\"o}rg Sander}, {and} \bibinfo{person}{Xiaowei
  Xu}.} \bibinfo{year}{1996}\natexlab{}.
\newblock \showarticletitle{A density-based algorithm for discovering clusters
  in large spatial databases with noise}.
\newblock \bibinfo{journal}{\emph{Proceedings of the 2nd International
  Conference on Knowledge Discovery and Data Mining (KDD)}}
  (\bibinfo{year}{1996}), \bibinfo{pages}{226--231}.
\newblock


\bibitem[Farahani et~al\mbox{.}(2021)]%
        {farahani2021parsbert}
\bibfield{author}{\bibinfo{person}{Mehrdad Farahani}, \bibinfo{person}{Mohammad
  Gharachorloo}, \bibinfo{person}{Marzieh Farahani}, {and}
  \bibinfo{person}{Mohammad Manthouri}.} \bibinfo{year}{2021}\natexlab{}.
\newblock \showarticletitle{{ParsBERT}: Transformer-based Model for {P}ersian
  Language Understanding}.
\newblock \bibinfo{journal}{\emph{Neural Processing Letters}}
  \bibinfo{volume}{53}, \bibinfo{number}{6} (\bibinfo{year}{2021}),
  \bibinfo{pages}{3831--3847}.
\newblock
\href{https://doi.org/10.1007/s11063-021-10528-4}{doi:\nolinkurl{10.1007/s11063-021-10528-4}}


\bibitem[Gao et~al\mbox{.}(2021)]%
        {gao2021advances}
\bibfield{author}{\bibinfo{person}{Chongming Gao}, \bibinfo{person}{Wenqiang
  Lei}, \bibinfo{person}{Xiangnan He}, \bibinfo{person}{Maarten de Rijke},
  {and} \bibinfo{person}{Tat-Seng Chua}.} \bibinfo{year}{2021}\natexlab{}.
\newblock \showarticletitle{Advances and Challenges in Conversational
  Recommender Systems: A Survey}.
\newblock \bibinfo{journal}{\emph{AI Open}}  \bibinfo{volume}{2}
  (\bibinfo{year}{2021}), \bibinfo{pages}{100--126}.
\newblock
\href{https://doi.org/10.1016/j.aiopen.2021.06.002}{doi:\nolinkurl{10.1016/j.aiopen.2021.06.002}}


\bibitem[{Gemini Team, Google}(2023)]%
        {gemini2023}
\bibfield{author}{\bibinfo{person}{{Gemini Team, Google}}.}
  \bibinfo{year}{2023}\natexlab{}.
\newblock \bibinfo{title}{Gemini: A Family of Highly Capable Multimodal
  Models}.
\newblock
\showeprint[arxiv]{2312.11805}
\urldef\tempurl%
\url{https://arxiv.org/abs/2312.11805}
\showURL{%
\tempurl}


\bibitem[{Gemma Team, Google DeepMind}(2025)]%
        {gemma2025}
\bibfield{author}{\bibinfo{person}{{Gemma Team, Google DeepMind}}.}
  \bibinfo{year}{2025}\natexlab{}.
\newblock \bibinfo{title}{Gemma 3 Technical Report}.
\newblock
\showeprint[arxiv]{2503.19786}
\urldef\tempurl%
\url{https://arxiv.org/abs/2503.19786}
\showURL{%
\tempurl}


\bibitem[Ghahroodi et~al\mbox{.}(2024)]%
        {ghahroodi2024khayyam}
\bibfield{author}{\bibinfo{person}{Omid Ghahroodi}, \bibinfo{person}{Marzia
  Nouri}, \bibinfo{person}{Mohammad~Vali Sanian}, \bibinfo{person}{Alireza
  Sahebi}, \bibinfo{person}{Doratossadat Dastgheib},
  \bibinfo{person}{Ehsaneddin Asgari}, \bibinfo{person}{Mahdieh
  Soleymani~Baghshah}, {and} \bibinfo{person}{Mohammad~Hossein Rohban}.}
  \bibinfo{year}{2024}\natexlab{}.
\newblock \bibinfo{title}{Khayyam Challenge ({PersianMMLU}): Is Your {LLM}
  Truly Wise to The {P}ersian Language?}
\newblock
\showeprint[arxiv]{2404.06644}~[cs.CL]
\urldef\tempurl%
\url{https://arxiv.org/abs/2404.06644}
\showURL{%
\tempurl}


\bibitem[Gu et~al\mbox{.}(2024)]%
        {gu2024survey}
\bibfield{author}{\bibinfo{person}{Jiawei Gu}, \bibinfo{person}{Xuhui Jiang},
  \bibinfo{person}{Zhichao Shi}, \bibinfo{person}{Hexiang Tan},
  \bibinfo{person}{Xuehao Zhai}, \bibinfo{person}{Chengjin Xu},
  \bibinfo{person}{Wei Li}, \bibinfo{person}{Yinghan Shen},
  \bibinfo{person}{Shengjie Ma}, \bibinfo{person}{Honghao Liu},
  \bibinfo{person}{Saizhuo Wang}, \bibinfo{person}{Kun Zhang},
  \bibinfo{person}{Yuanzhuo Wang}, \bibinfo{person}{Wen Gao},
  \bibinfo{person}{Lionel Ni}, {and} \bibinfo{person}{Jian Guo}.}
  \bibinfo{year}{2024}\natexlab{}.
\newblock \bibinfo{title}{A Survey on {LLM}-as-a-Judge}.
\newblock
\showeprint[arxiv]{2411.15594}~[cs.CL]
\urldef\tempurl%
\url{https://arxiv.org/abs/2411.15594}
\showURL{%
\tempurl}


\bibitem[Hu et~al\mbox{.}(2024)]%
        {hu2024uncertainty}
\bibfield{author}{\bibinfo{person}{Zhiyuan Hu}, \bibinfo{person}{Chumin Liu},
  \bibinfo{person}{Xidong Feng}, \bibinfo{person}{Yilun Zhao},
  \bibinfo{person}{See-Kiong Ng}, \bibinfo{person}{Anh~Tuan Luu},
  \bibinfo{person}{Junxian He}, \bibinfo{person}{Pang~Wei Koh}, {and}
  \bibinfo{person}{Bryan Hooi}.} \bibinfo{year}{2024}\natexlab{}.
\newblock \showarticletitle{Uncertainty of Thoughts: Uncertainty-Aware Planning
  Enhances Information Seeking in {LLM}s}. In
  \bibinfo{booktitle}{\emph{Advances in Neural Information Processing Systems
  37 (NeurIPS 2024)}}. \bibinfo{pages}{24181--24215}.
\newblock
\href{https://doi.org/10.52202/079017-0762}{doi:\nolinkurl{10.52202/079017-0762}}


\bibitem[Jannach et~al\mbox{.}(2021)]%
        {jannach2021survey}
\bibfield{author}{\bibinfo{person}{Dietmar Jannach}, \bibinfo{person}{Ahtsham
  Manzoor}, \bibinfo{person}{Wanling Cai}, {and} \bibinfo{person}{Li Chen}.}
  \bibinfo{year}{2021}\natexlab{}.
\newblock \showarticletitle{A Survey on Conversational Recommender Systems}.
\newblock \bibinfo{journal}{\emph{Comput. Surveys}} \bibinfo{volume}{54},
  \bibinfo{number}{5} (\bibinfo{year}{2021}), \bibinfo{pages}{105:1--105:36}.
\newblock
\href{https://doi.org/10.1145/3453154}{doi:\nolinkurl{10.1145/3453154}}


\bibitem[Khashabi et~al\mbox{.}(2021)]%
        {khashabi2021parsinlu}
\bibfield{author}{\bibinfo{person}{Daniel Khashabi}, \bibinfo{person}{Arman
  Cohan}, \bibinfo{person}{Siamak Shakeri}, \bibinfo{person}{Pedram Hosseini},
  \bibinfo{person}{Pouya Pezeshkpour}, \bibinfo{person}{Malihe Alikhani},
  \bibinfo{person}{Moin Aminnaseri}, \bibinfo{person}{Marzieh Bitaab},
  \bibinfo{person}{Faeze Brahman}, \bibinfo{person}{Sarik Ghazarian},
  {et~al\mbox{.}}} \bibinfo{year}{2021}\natexlab{}.
\newblock \showarticletitle{{ParsiNLU}: A Suite of Language Understanding
  Challenges for {P}ersian}.
\newblock \bibinfo{journal}{\emph{Transactions of the Association for
  Computational Linguistics}}  \bibinfo{volume}{9} (\bibinfo{year}{2021}),
  \bibinfo{pages}{1147--1162}.
\newblock
\href{https://doi.org/10.1162/tacl_a_00419}{doi:\nolinkurl{10.1162/tacl_a_00419}}


\bibitem[Kim et~al\mbox{.}(2023)]%
        {kim2023soda}
\bibfield{author}{\bibinfo{person}{Hyunwoo Kim}, \bibinfo{person}{Jack Hessel},
  \bibinfo{person}{Liwei Jiang}, \bibinfo{person}{Peter West},
  \bibinfo{person}{Ximing Lu}, \bibinfo{person}{Youngjae Yu},
  \bibinfo{person}{Pei Zhou}, \bibinfo{person}{Ronan~Le Bras},
  \bibinfo{person}{Malihe Alikhani}, \bibinfo{person}{Gunhee Kim},
  \bibinfo{person}{Maarten Sap}, {and} \bibinfo{person}{Yejin Choi}.}
  \bibinfo{year}{2023}\natexlab{}.
\newblock \showarticletitle{{SODA}: Million-scale Dialogue Distillation with
  Social Commonsense Contextualization}. In
  \bibinfo{booktitle}{\emph{Proceedings of the 2023 Conference on Empirical
  Methods in Natural Language Processing (EMNLP)}}.
  \bibinfo{pages}{12930--12949}.
\newblock
\href{https://doi.org/10.18653/v1/2023.emnlp-main.799}{doi:\nolinkurl{10.18653/v1/2023.emnlp-main.799}}


\bibitem[Kim et~al\mbox{.}(2024)]%
        {kim2024pearl}
\bibfield{author}{\bibinfo{person}{Minjin Kim}, \bibinfo{person}{Minju Kim},
  \bibinfo{person}{Hana Kim}, \bibinfo{person}{Beong-woo Kwak},
  \bibinfo{person}{SeongKu Kang}, \bibinfo{person}{Youngjae Yu},
  \bibinfo{person}{Jinyoung Yeo}, {and} \bibinfo{person}{Dongha Lee}.}
  \bibinfo{year}{2024}\natexlab{}.
\newblock \showarticletitle{Pearl: A Review-driven Persona-Knowledge Grounded
  Conversational Recommendation Dataset}. In \bibinfo{booktitle}{\emph{Findings
  of the Association for Computational Linguistics: ACL 2024}},
  \bibfield{editor}{\bibinfo{person}{Lun-Wei Ku}, \bibinfo{person}{Andre
  Martins}, {and} \bibinfo{person}{Vivek Srikumar}} (Eds.).
  \bibinfo{publisher}{Association for Computational Linguistics},
  \bibinfo{address}{Bangkok, Thailand}, \bibinfo{pages}{1105--1120}.
\newblock
\href{https://doi.org/10.18653/v1/2024.findings-acl.65}{doi:\nolinkurl{10.18653/v1/2024.findings-acl.65}}


\bibitem[Kish(1965)]%
        {kish1965survey}
\bibfield{author}{\bibinfo{person}{Leslie Kish}.}
  \bibinfo{year}{1965}\natexlab{}.
\newblock \bibinfo{booktitle}{\emph{Survey Sampling}}.
\newblock \bibinfo{publisher}{John Wiley \& Sons}, \bibinfo{address}{New York}.
\newblock


\bibitem[Lei et~al\mbox{.}(2020a)]%
        {lei2020estimation}
\bibfield{author}{\bibinfo{person}{Wenqiang Lei}, \bibinfo{person}{Xiangnan
  He}, \bibinfo{person}{Yisong Miao}, \bibinfo{person}{Qingyun Wu},
  \bibinfo{person}{Richang Hong}, \bibinfo{person}{Min-Yen Kan}, {and}
  \bibinfo{person}{Tat-Seng Chua}.} \bibinfo{year}{2020}\natexlab{a}.
\newblock \showarticletitle{Estimation-Action-Reflection: Towards Deep
  Interaction Between Conversational and Recommender Systems}. In
  \bibinfo{booktitle}{\emph{Proceedings of the 13th International Conference on
  Web Search and Data Mining}}. \bibinfo{pages}{304--312}.
\newblock
\href{https://doi.org/10.1145/3336191.3371769}{doi:\nolinkurl{10.1145/3336191.3371769}}


\bibitem[Lei et~al\mbox{.}(2020b)]%
        {lei2020interactive}
\bibfield{author}{\bibinfo{person}{Wenqiang Lei}, \bibinfo{person}{Gangyi
  Zhang}, \bibinfo{person}{Xiangnan He}, \bibinfo{person}{Yisong Miao},
  \bibinfo{person}{Xiang Wang}, \bibinfo{person}{Liang Chen}, {and}
  \bibinfo{person}{Tat-Seng Chua}.} \bibinfo{year}{2020}\natexlab{b}.
\newblock \showarticletitle{Interactive Path Reasoning on Graph for
  Conversational Recommendation}. In \bibinfo{booktitle}{\emph{Proceedings of
  the 26th ACM SIGKDD International Conference on Knowledge Discovery and Data
  Mining}}. \bibinfo{pages}{2073--2083}.
\newblock
\href{https://doi.org/10.1145/3394486.3403258}{doi:\nolinkurl{10.1145/3394486.3403258}}


\bibitem[Leszczynski et~al\mbox{.}(2023)]%
        {leszczynski2023talk}
\bibfield{author}{\bibinfo{person}{Megan Leszczynski}, \bibinfo{person}{Shu
  Zhang}, \bibinfo{person}{Ravi Ganti}, \bibinfo{person}{Krisztian Balog},
  \bibinfo{person}{Filip Radlinski}, \bibinfo{person}{Fernando Pereira}, {and}
  \bibinfo{person}{Arun~Tejasvi Chaganty}.} \bibinfo{year}{2023}\natexlab{}.
\newblock \bibinfo{title}{Talk the Walk: Synthetic Data Generation for
  Conversational Music Recommendation}.
\newblock
\showeprint[arxiv]{2301.11489}~[cs.IR]
\urldef\tempurl%
\url{https://arxiv.org/abs/2301.11489}
\showURL{%
\tempurl}


\bibitem[Li et~al\mbox{.}(2018)]%
        {li2018towards}
\bibfield{author}{\bibinfo{person}{Raymond Li},
  \bibinfo{person}{Samira~Ebrahimi Kahou}, \bibinfo{person}{Hannes Schulz},
  \bibinfo{person}{Vincent Michalski}, \bibinfo{person}{Laurent Charlin}, {and}
  \bibinfo{person}{Chris Pal}.} \bibinfo{year}{2018}\natexlab{}.
\newblock \showarticletitle{Towards Deep Conversational Recommendations}. In
  \bibinfo{booktitle}{\emph{Advances in Neural Information Processing Systems
  31 (NeurIPS)}}. \bibinfo{pages}{9748--9758}.
\newblock


\bibitem[Li et~al\mbox{.}(2025)]%
        {li2025wizardshoppingtargetorientedecommerce}
\bibfield{author}{\bibinfo{person}{Xiangci Li}, \bibinfo{person}{Zhiyu Chen},
  \bibinfo{person}{Jason~Ingyu Choi}, \bibinfo{person}{Nikhita Vedula},
  \bibinfo{person}{Besnik Fetahu}, \bibinfo{person}{Oleg Rokhlenko}, {and}
  \bibinfo{person}{Shervin Malmasi}.} \bibinfo{year}{2025}\natexlab{}.
\newblock \bibinfo{title}{Wizard of Shopping: Target-Oriented E-commerce
  Dialogue Generation with Decision Tree Branching}.
\newblock
\showeprint[arxiv]{2502.00969}~[cs.CL]
\urldef\tempurl%
\url{https://arxiv.org/abs/2502.00969}
\showURL{%
\tempurl}


\bibitem[Liang et~al\mbox{.}(2024)]%
        {liang2024llmredial}
\bibfield{author}{\bibinfo{person}{Tingting Liang}, \bibinfo{person}{Chenxin
  Jin}, \bibinfo{person}{Lingzhi Wang}, \bibinfo{person}{Wenqi Fan},
  \bibinfo{person}{Congying Xia}, \bibinfo{person}{Kai Chen}, {and}
  \bibinfo{person}{Yuyu Yin}.} \bibinfo{year}{2024}\natexlab{}.
\newblock \showarticletitle{{LLM}-{REDIAL}: A Large-Scale Dataset for
  Conversational Recommender Systems Created from User Behaviors with {LLM}s}.
  In \bibinfo{booktitle}{\emph{Findings of the Association for Computational
  Linguistics: ACL 2024}}, \bibfield{editor}{\bibinfo{person}{Lun-Wei Ku},
  \bibinfo{person}{Andre Martins}, {and} \bibinfo{person}{Vivek Srikumar}}
  (Eds.). \bibinfo{publisher}{Association for Computational Linguistics},
  \bibinfo{address}{Bangkok, Thailand}, \bibinfo{pages}{8926--8939}.
\newblock
\href{https://doi.org/10.18653/v1/2024.findings-acl.529}{doi:\nolinkurl{10.18653/v1/2024.findings-acl.529}}


\bibitem[Liu et~al\mbox{.}(2023)]%
        {liu2023geval}
\bibfield{author}{\bibinfo{person}{Yang Liu}, \bibinfo{person}{Dan Iter},
  \bibinfo{person}{Yichong Xu}, \bibinfo{person}{Shuohang Wang},
  \bibinfo{person}{Ruochen Xu}, {and} \bibinfo{person}{Chenguang Zhu}.}
  \bibinfo{year}{2023}\natexlab{}.
\newblock \showarticletitle{{G-Eval}: {NLG} Evaluation using {GPT}-4 with
  Better Human Alignment}. In \bibinfo{booktitle}{\emph{Proceedings of the 2023
  Conference on Empirical Methods in Natural Language Processing (EMNLP)}}.
  \bibinfo{pages}{2511--2522}.
\newblock
\href{https://doi.org/10.18653/v1/2023.emnlp-main.153}{doi:\nolinkurl{10.18653/v1/2023.emnlp-main.153}}


\bibitem[Long et~al\mbox{.}(2024)]%
        {long2024synthetic}
\bibfield{author}{\bibinfo{person}{Lin Long}, \bibinfo{person}{Rui Wang},
  \bibinfo{person}{Ruixuan Xiao}, \bibinfo{person}{Junbo Zhao},
  \bibinfo{person}{Xiao Ding}, \bibinfo{person}{Gang Chen}, {and}
  \bibinfo{person}{Haobo Wang}.} \bibinfo{year}{2024}\natexlab{}.
\newblock \showarticletitle{On {LLM}s-Driven Synthetic Data Generation,
  Curation, and Evaluation: A Survey}. In \bibinfo{booktitle}{\emph{Findings of
  the Association for Computational Linguistics: ACL 2024}},
  \bibfield{editor}{\bibinfo{person}{Lun-Wei Ku}, \bibinfo{person}{Andre
  Martins}, {and} \bibinfo{person}{Vivek Srikumar}} (Eds.).
  \bibinfo{publisher}{Association for Computational Linguistics},
  \bibinfo{address}{Bangkok, Thailand}, \bibinfo{pages}{11065--11082}.
\newblock
\href{https://doi.org/10.18653/v1/2024.findings-acl.658}{doi:\nolinkurl{10.18653/v1/2024.findings-acl.658}}


\bibitem[MacQueen(1967)]%
        {macqueen1967some}
\bibfield{author}{\bibinfo{person}{J. MacQueen}.}
  \bibinfo{year}{1967}\natexlab{}.
\newblock \showarticletitle{Some methods for classification and analysis of
  multivariate observations}. In \bibinfo{booktitle}{\emph{Proceedings of the
  fifth Berkeley symposium on mathematical statistics and probability}},
  Vol.~\bibinfo{volume}{1}. Oakland, CA, USA, \bibinfo{pages}{281--297}.
\newblock


\bibitem[Mahmoudi and Faili(2024)]%
        {mahmoudi2024pershoppersiandataset}
\bibfield{author}{\bibinfo{person}{Keyvan Mahmoudi} {and}
  \bibinfo{person}{Heshaam Faili}.} \bibinfo{year}{2024}\natexlab{}.
\newblock \bibinfo{title}{PerSHOP -- A Persian dataset for shopping dialogue
  systems modeling}.
\newblock
\showeprint[arxiv]{2401.00811}~[cs.CL]
\urldef\tempurl%
\url{https://arxiv.org/abs/2401.00811}
\showURL{%
\tempurl}


\bibitem[Montazeralghaem and Allan(2022)]%
        {montazeralghaem2022learning}
\bibfield{author}{\bibinfo{person}{Ali Montazeralghaem} {and}
  \bibinfo{person}{James Allan}.} \bibinfo{year}{2022}\natexlab{}.
\newblock \showarticletitle{Learning Relevant Questions for Conversational
  Product Search using Deep Reinforcement Learning}. In
  \bibinfo{booktitle}{\emph{Proceedings of the Fifteenth ACM International
  Conference on Web Search and Data Mining}} \emph{(\bibinfo{series}{WSDM
  '22})}. \bibinfo{publisher}{ACM}, \bibinfo{pages}{746--754}.
\newblock
\href{https://doi.org/10.1145/3488560.3498526}{doi:\nolinkurl{10.1145/3488560.3498526}}


\bibitem[{OpenAI}(2024)]%
        {openai2024gpt4o}
\bibfield{author}{\bibinfo{person}{{OpenAI}}.} \bibinfo{year}{2024}\natexlab{}.
\newblock \bibinfo{title}{{GPT-4o} System Card}.
\newblock
\showeprint[arxiv]{2410.21276}
\urldef\tempurl%
\url{https://arxiv.org/abs/2410.21276}
\showURL{%
\tempurl}


\bibitem[Panickssery et~al\mbox{.}(2024)]%
        {panickssery2024llm}
\bibfield{author}{\bibinfo{person}{Arjun Panickssery},
  \bibinfo{person}{Samuel~R. Bowman}, {and} \bibinfo{person}{Shi Feng}.}
  \bibinfo{year}{2024}\natexlab{}.
\newblock \showarticletitle{{LLM} Evaluators Recognize and Favor Their Own
  Generations}.
\newblock \bibinfo{journal}{\emph{Advances in Neural Information Processing
  Systems}}  \bibinfo{volume}{37} (\bibinfo{year}{2024}).
\newblock


\bibitem[Quinlan(1986)]%
        {quinlan1986induction}
\bibfield{author}{\bibinfo{person}{J.~Ross Quinlan}.}
  \bibinfo{year}{1986}\natexlab{}.
\newblock \showarticletitle{Induction of Decision Trees}.
\newblock \bibinfo{journal}{\emph{Machine Learning}} \bibinfo{volume}{1},
  \bibinfo{number}{1} (\bibinfo{year}{1986}), \bibinfo{pages}{81--106}.
\newblock
\href{https://doi.org/10.1007/BF00116251}{doi:\nolinkurl{10.1007/BF00116251}}


\bibitem[Radlinski and Craswell(2017)]%
        {radlinski2017theoretical}
\bibfield{author}{\bibinfo{person}{Filip Radlinski} {and} \bibinfo{person}{Nick
  Craswell}.} \bibinfo{year}{2017}\natexlab{}.
\newblock \showarticletitle{A Theoretical Framework for Conversational Search}.
  In \bibinfo{booktitle}{\emph{Proceedings of the 2017 Conference on Human
  Information Interaction and Retrieval (CHIIR)}}. \bibinfo{pages}{117--126}.
\newblock
\href{https://doi.org/10.1145/3020165.3020183}{doi:\nolinkurl{10.1145/3020165.3020183}}


\bibitem[Rahmani et~al\mbox{.}(2025)]%
        {rahmani2025megachat}
\bibfield{author}{\bibinfo{person}{Mahdi Rahmani}, \bibinfo{person}{AmirHossein
  Saffari}, {and} \bibinfo{person}{Reyhane Rahmani}.}
  \bibinfo{year}{2025}\natexlab{}.
\newblock \bibinfo{title}{{MegaChat}: A Synthetic {P}ersian {Q\&A} Dataset for
  High-Quality Sales Chatbot Evaluation}.
\newblock
\showeprint[arxiv]{2511.23397}
\urldef\tempurl%
\url{https://arxiv.org/abs/2511.23397}
\showURL{%
\tempurl}


\bibitem[Reddy et~al\mbox{.}(2022)]%
        {reddy2022shopping}
\bibfield{author}{\bibinfo{person}{Chandan~K. Reddy},
  \bibinfo{person}{Llu{\'i}s M{\`a}rquez}, \bibinfo{person}{Fran Valero},
  \bibinfo{person}{Nikhil Rao}, \bibinfo{person}{Hugo Zaragoza},
  \bibinfo{person}{Sambaran Bandyopadhyay}, \bibinfo{person}{Arnab Biswas},
  \bibinfo{person}{Anlu Xing}, {and} \bibinfo{person}{Karthik Subbian}.}
  \bibinfo{year}{2022}\natexlab{}.
\newblock \bibinfo{title}{Shopping Queries Dataset: A Large-Scale {ESCI}
  Benchmark for Improving Product Search}.
\newblock
\showeprint[arxiv]{2206.06588}~[cs.IR]
\urldef\tempurl%
\url{https://arxiv.org/abs/2206.06588}
\showURL{%
\tempurl}


\bibitem[Reimers and Gurevych(2019)]%
        {reimers2019sentence}
\bibfield{author}{\bibinfo{person}{Nils Reimers} {and} \bibinfo{person}{Iryna
  Gurevych}.} \bibinfo{year}{2019}\natexlab{}.
\newblock \showarticletitle{Sentence-{BERT}: Sentence Embeddings using
  {S}iamese {BERT}-Networks}. In \bibinfo{booktitle}{\emph{Proceedings of the
  2019 Conference on Empirical Methods in Natural Language Processing and the
  9th International Joint Conference on Natural Language Processing
  (EMNLP-IJCNLP)}}. \bibinfo{pages}{3982--3992}.
\newblock
\href{https://doi.org/10.18653/v1/D19-1410}{doi:\nolinkurl{10.18653/v1/D19-1410}}


\bibitem[Reimers and Gurevych(2020)]%
        {reimers2020making}
\bibfield{author}{\bibinfo{person}{Nils Reimers} {and} \bibinfo{person}{Iryna
  Gurevych}.} \bibinfo{year}{2020}\natexlab{}.
\newblock \showarticletitle{Making Monolingual Sentence Embeddings Multilingual
  using Knowledge Distillation}. In \bibinfo{booktitle}{\emph{Proceedings of
  the 2020 Conference on Empirical Methods in Natural Language Processing
  (EMNLP)}}. \bibinfo{pages}{4512--4525}.
\newblock
\href{https://doi.org/10.18653/v1/2020.emnlp-main.365}{doi:\nolinkurl{10.18653/v1/2020.emnlp-main.365}}


\bibitem[Rousseeuw(1987)]%
        {rousseeuw1987silhouettes}
\bibfield{author}{\bibinfo{person}{Peter~J Rousseeuw}.}
  \bibinfo{year}{1987}\natexlab{}.
\newblock \showarticletitle{Silhouettes: a graphical aid to the interpretation
  and validation of cluster analysis}.
\newblock \bibinfo{journal}{\emph{Journal of computational and applied
  mathematics}}  \bibinfo{volume}{20} (\bibinfo{year}{1987}),
  \bibinfo{pages}{53--65}.
\newblock


\bibitem[Salton and Buckley(1988)]%
        {salton1988term}
\bibfield{author}{\bibinfo{person}{Gerard Salton} {and}
  \bibinfo{person}{Christopher Buckley}.} \bibinfo{year}{1988}\natexlab{}.
\newblock \showarticletitle{Term-weighting approaches in automatic text
  retrieval}.
\newblock \bibinfo{journal}{\emph{Information processing \& management}}
  \bibinfo{volume}{24}, \bibinfo{number}{5} (\bibinfo{year}{1988}),
  \bibinfo{pages}{513--523}.
\newblock


\bibitem[Shim et~al\mbox{.}(2025)]%
        {shim2025tooldialmultiturndialoguegeneration}
\bibfield{author}{\bibinfo{person}{Jeonghoon Shim}, \bibinfo{person}{Gyuhyeon
  Seo}, \bibinfo{person}{Cheongsu Lim}, {and} \bibinfo{person}{Yohan Jo}.}
  \bibinfo{year}{2025}\natexlab{}.
\newblock \bibinfo{title}{ToolDial: Multi-turn Dialogue Generation Method for
  Tool-Augmented Language Models}.
\newblock
\showeprint[arxiv]{2503.00564}~[cs.CL]
\urldef\tempurl%
\url{https://arxiv.org/abs/2503.00564}
\showURL{%
\tempurl}


\bibitem[Sun and Zhang(2018)]%
        {sun2018conversational}
\bibfield{author}{\bibinfo{person}{Yueming Sun} {and} \bibinfo{person}{Yi
  Zhang}.} \bibinfo{year}{2018}\natexlab{}.
\newblock \showarticletitle{Conversational Recommender System}. In
  \bibinfo{booktitle}{\emph{Proceedings of the 41st International ACM SIGIR
  Conference on Research and Development in Information Retrieval}}.
  \bibinfo{pages}{235--244}.
\newblock
\href{https://doi.org/10.1145/3209978.3210002}{doi:\nolinkurl{10.1145/3209978.3210002}}


\bibitem[Suresh et~al\mbox{.}(2025)]%
        {suresh-etal-2025-diasynth}
\bibfield{author}{\bibinfo{person}{Sathya~Krishnan Suresh}, \bibinfo{person}{Wu
  Mengjun}, \bibinfo{person}{Tushar Pranav}, {and} \bibinfo{person}{EngSiong
  Chng}.} \bibinfo{year}{2025}\natexlab{}.
\newblock \showarticletitle{{D}ia{S}ynth: Synthetic Dialogue Generation
  Framework for Low Resource Dialogue Applications}. In
  \bibinfo{booktitle}{\emph{Findings of the Association for Computational
  Linguistics: NAACL 2025}}, \bibfield{editor}{\bibinfo{person}{Luis Chiruzzo},
  \bibinfo{person}{Alan Ritter}, {and} \bibinfo{person}{Lu~Wang}} (Eds.).
  \bibinfo{publisher}{Association for Computational Linguistics},
  \bibinfo{address}{Albuquerque, New Mexico}, \bibinfo{pages}{673--690}.
\newblock
\showISBNx{979-8-89176-195-7}
\href{https://doi.org/10.18653/v1/2025.findings-naacl.40}{doi:\nolinkurl{10.18653/v1/2025.findings-naacl.40}}


\bibitem[Wang et~al\mbox{.}(2025)]%
        {wang2025shoppingbenchrealworldintentgroundedshopping}
\bibfield{author}{\bibinfo{person}{Jiangyuan Wang}, \bibinfo{person}{Kejun
  Xiao}, \bibinfo{person}{Qi Sun}, \bibinfo{person}{Huaipeng Zhao},
  \bibinfo{person}{Tao Luo}, \bibinfo{person}{Jian~Dong Zhang}, {and}
  \bibinfo{person}{Xiaoyi Zeng}.} \bibinfo{year}{2025}\natexlab{}.
\newblock \bibinfo{title}{ShoppingBench: A Real-World Intent-Grounded Shopping
  Benchmark for LLM-based Agents}.
\newblock
\showeprint[arxiv]{2508.04266}~[cs.CL]
\urldef\tempurl%
\url{https://arxiv.org/abs/2508.04266}
\showURL{%
\tempurl}


\bibitem[Wang et~al\mbox{.}(2024)]%
        {wang2024fair}
\bibfield{author}{\bibinfo{person}{Peiyi Wang}, \bibinfo{person}{Lei Li},
  \bibinfo{person}{Liang Chen}, \bibinfo{person}{Zefan Cai},
  \bibinfo{person}{Dawei Zhu}, \bibinfo{person}{Binghuai Lin},
  \bibinfo{person}{Yunbo Cao}, \bibinfo{person}{Lingpeng Kong},
  \bibinfo{person}{Qi Liu}, \bibinfo{person}{Tianyu Liu}, {and}
  \bibinfo{person}{Zhifang Sui}.} \bibinfo{year}{2024}\natexlab{}.
\newblock \showarticletitle{Large Language Models are not Fair Evaluators}. In
  \bibinfo{booktitle}{\emph{Proceedings of the 62nd Annual Meeting of the
  Association for Computational Linguistics (Volume 1: Long Papers)}},
  \bibfield{editor}{\bibinfo{person}{Lun-Wei Ku}, \bibinfo{person}{Andre
  Martins}, {and} \bibinfo{person}{Vivek Srikumar}} (Eds.).
  \bibinfo{publisher}{Association for Computational Linguistics},
  \bibinfo{address}{Bangkok, Thailand}, \bibinfo{pages}{9440--9450}.
\newblock
\href{https://doi.org/10.18653/v1/2024.acl-long.511}{doi:\nolinkurl{10.18653/v1/2024.acl-long.511}}


\bibitem[Wang et~al\mbox{.}(2020)]%
        {10.5555/3495724.3496209}
\bibfield{author}{\bibinfo{person}{Wenhui Wang}, \bibinfo{person}{Furu Wei},
  \bibinfo{person}{Li Dong}, \bibinfo{person}{Hangbo Bao}, \bibinfo{person}{Nan
  Yang}, {and} \bibinfo{person}{Ming Zhou}.} \bibinfo{year}{2020}\natexlab{}.
\newblock \showarticletitle{MINILM: deep self-attention distillation for
  task-agnostic compression of pre-trained transformers}. In
  \bibinfo{booktitle}{\emph{Proceedings of the 34th International Conference on
  Neural Information Processing Systems}} (Vancouver, BC, Canada)
  \emph{(\bibinfo{series}{NIPS '20})}. \bibinfo{publisher}{Curran Associates
  Inc.}, \bibinfo{address}{Red Hook, NY, USA}, Article
  \bibinfo{articleno}{485}, \bibinfo{numpages}{13}~pages.
\newblock
\showISBNx{9781713829546}


\bibitem[Wang et~al\mbox{.}(2023)]%
        {wang2023rethinking}
\bibfield{author}{\bibinfo{person}{Xiaolei Wang}, \bibinfo{person}{Xinyu Tang},
  \bibinfo{person}{Wayne~Xin Zhao}, \bibinfo{person}{Jingyuan Wang}, {and}
  \bibinfo{person}{Ji-Rong Wen}.} \bibinfo{year}{2023}\natexlab{}.
\newblock \showarticletitle{Rethinking the Evaluation for Conversational
  Recommendation in the Era of Large Language Models}. In
  \bibinfo{booktitle}{\emph{Proceedings of the 2023 Conference on Empirical
  Methods in Natural Language Processing}},
  \bibfield{editor}{\bibinfo{person}{Houda Bouamor}, \bibinfo{person}{Juan
  Pino}, {and} \bibinfo{person}{Kalika Bali}} (Eds.).
  \bibinfo{publisher}{Association for Computational Linguistics},
  \bibinfo{address}{Singapore}, \bibinfo{pages}{10052--10065}.
\newblock
\href{https://doi.org/10.18653/v1/2023.emnlp-main.621}{doi:\nolinkurl{10.18653/v1/2023.emnlp-main.621}}


\bibitem[Yoon et~al\mbox{.}(2024)]%
        {yoon2024evaluating}
\bibfield{author}{\bibinfo{person}{Se-eun Yoon}, \bibinfo{person}{Zhankui He},
  \bibinfo{person}{Jessica Echterhoff}, {and} \bibinfo{person}{Julian
  McAuley}.} \bibinfo{year}{2024}\natexlab{}.
\newblock \showarticletitle{Evaluating Large Language Models as Generative User
  Simulators for Conversational Recommendation}. In
  \bibinfo{booktitle}{\emph{Proceedings of the 2024 Conference of the North
  American Chapter of the Association for Computational Linguistics: Human
  Language Technologies (Volume 1: Long Papers)}},
  \bibfield{editor}{\bibinfo{person}{Kevin Duh}, \bibinfo{person}{Helena
  Gomez}, {and} \bibinfo{person}{Steven Bethard}} (Eds.).
  \bibinfo{publisher}{Association for Computational Linguistics},
  \bibinfo{address}{Mexico City, Mexico}, \bibinfo{pages}{1490--1504}.
\newblock
\href{https://doi.org/10.18653/v1/2024.naacl-long.83}{doi:\nolinkurl{10.18653/v1/2024.naacl-long.83}}


\bibitem[Zamani et~al\mbox{.}(2023)]%
        {zamani2023conversational}
\bibfield{author}{\bibinfo{person}{Hamed Zamani}, \bibinfo{person}{Johanne~R.
  Trippas}, \bibinfo{person}{Jeff Dalton}, {and} \bibinfo{person}{Filip
  Radlinski}.} \bibinfo{year}{2023}\natexlab{}.
\newblock \showarticletitle{Conversational Information Seeking}.
\newblock \bibinfo{journal}{\emph{Foundations and Trends in Information
  Retrieval}} \bibinfo{volume}{17}, \bibinfo{number}{3--4}
  (\bibinfo{year}{2023}), \bibinfo{pages}{244--456}.
\newblock
\href{https://doi.org/10.1561/1500000081}{doi:\nolinkurl{10.1561/1500000081}}


\bibitem[Zhang and Balog(2020)]%
        {zhang2020evaluating}
\bibfield{author}{\bibinfo{person}{Shuo Zhang} {and} \bibinfo{person}{Krisztian
  Balog}.} \bibinfo{year}{2020}\natexlab{}.
\newblock \showarticletitle{Evaluating Conversational Recommender Systems via
  User Simulation}. In \bibinfo{booktitle}{\emph{Proceedings of the 26th ACM
  SIGKDD International Conference on Knowledge Discovery \& Data Mining}}
  \emph{(\bibinfo{series}{KDD '20})}. \bibinfo{publisher}{ACM},
  \bibinfo{pages}{1512--1520}.
\newblock
\href{https://doi.org/10.1145/3394486.3403202}{doi:\nolinkurl{10.1145/3394486.3403202}}


\bibitem[Zhang et~al\mbox{.}(2018)]%
        {zhang2018towards}
\bibfield{author}{\bibinfo{person}{Yongfeng Zhang}, \bibinfo{person}{Xu Chen},
  \bibinfo{person}{Qingyao Ai}, \bibinfo{person}{Liu Yang}, {and}
  \bibinfo{person}{W.~Bruce Croft}.} \bibinfo{year}{2018}\natexlab{}.
\newblock \showarticletitle{Towards Conversational Search and Recommendation:
  System Ask, User Respond}. In \bibinfo{booktitle}{\emph{Proceedings of the
  27th ACM International Conference on Information and Knowledge Management
  (CIKM)}}. \bibinfo{pages}{177--186}.
\newblock
\href{https://doi.org/10.1145/3269206.3271776}{doi:\nolinkurl{10.1145/3269206.3271776}}


\bibitem[Zheng et~al\mbox{.}(2023)]%
        {zheng2023judging}
\bibfield{author}{\bibinfo{person}{Lianmin Zheng}, \bibinfo{person}{Wei-Lin
  Chiang}, \bibinfo{person}{Ying Sheng}, \bibinfo{person}{Siyuan Zhuang},
  \bibinfo{person}{Zhanghao Wu}, \bibinfo{person}{Yonghao Zhuang},
  \bibinfo{person}{Zi Lin}, \bibinfo{person}{Zhuohan Li},
  \bibinfo{person}{Dacheng Li}, \bibinfo{person}{Eric~P. Xing},
  \bibinfo{person}{Hao Zhang}, \bibinfo{person}{Joseph~E. Gonzalez}, {and}
  \bibinfo{person}{Ion Stoica}.} \bibinfo{year}{2023}\natexlab{}.
\newblock \showarticletitle{Judging {LLM}-as-a-Judge with {MT}-Bench and
  Chatbot Arena}. In \bibinfo{booktitle}{\emph{Advances in Neural Information
  Processing Systems 36 (NeurIPS), Datasets and Benchmarks Track}}.
\newblock


\bibitem[Zhu et~al\mbox{.}(2024)]%
        {zhu2024reliable}
\bibfield{author}{\bibinfo{person}{Lixi Zhu}, \bibinfo{person}{Xiaowen Huang},
  {and} \bibinfo{person}{Jitao Sang}.} \bibinfo{year}{2024}\natexlab{}.
\newblock \showarticletitle{How Reliable is Your Simulator? {A}nalysis on the
  Limitations of Current {LLM}-based User Simulators for Conversational
  Recommendation}. In \bibinfo{booktitle}{\emph{Companion Proceedings of the
  ACM Web Conference 2024}} \emph{(\bibinfo{series}{WWW '24})}.
  \bibinfo{publisher}{ACM}, \bibinfo{pages}{1726--1732}.
\newblock
\href{https://doi.org/10.1145/3589335.3651955}{doi:\nolinkurl{10.1145/3589335.3651955}}


\bibitem[Zinvandi et~al\mbox{.}(2025)]%
        {zinvandi2025famteb}
\bibfield{author}{\bibinfo{person}{Erfan Zinvandi}, \bibinfo{person}{Morteza
  Alikhani}, \bibinfo{person}{Mehran Sarmadi}, \bibinfo{person}{Zahra
  Pourbahman}, \bibinfo{person}{Sepehr Arvin}, \bibinfo{person}{Reza Kazemi},
  {and} \bibinfo{person}{Arash Amini}.} \bibinfo{year}{2025}\natexlab{}.
\newblock \bibinfo{title}{{FaMTEB}: Massive Text Embedding Benchmark in
  {P}ersian Language}.
\newblock
\showeprint[arxiv]{2502.11571}
\urldef\tempurl%
\url{https://arxiv.org/abs/2502.11571}
\showURL{%
\tempurl}


\bibitem[Zou et~al\mbox{.}(2020)]%
        {zou2020towards}
\bibfield{author}{\bibinfo{person}{Jie Zou}, \bibinfo{person}{Yifan Chen},
  {and} \bibinfo{person}{Evangelos Kanoulas}.} \bibinfo{year}{2020}\natexlab{}.
\newblock \showarticletitle{Towards Question-Based Recommender Systems}. In
  \bibinfo{booktitle}{\emph{Proceedings of the 43rd International ACM SIGIR
  Conference on Research and Development in Information Retrieval}}.
  \bibinfo{pages}{881--890}.
\newblock
\href{https://doi.org/10.1145/3397271.3401180}{doi:\nolinkurl{10.1145/3397271.3401180}}


\bibitem[Zou et~al\mbox{.}(2022)]%
        {zou2022learning}
\bibfield{author}{\bibinfo{person}{Jie Zou}, \bibinfo{person}{Jimmy Huang},
  \bibinfo{person}{Zhaochun Ren}, {and} \bibinfo{person}{Evangelos Kanoulas}.}
  \bibinfo{year}{2022}\natexlab{}.
\newblock \showarticletitle{Learning to Ask: Conversational Product Search via
  Representation Learning}.
\newblock \bibinfo{journal}{\emph{ACM Transactions on Information Systems}}
  \bibinfo{volume}{41}, \bibinfo{number}{2} (\bibinfo{year}{2022}),
  \bibinfo{pages}{1--27}.
\newblock
\href{https://doi.org/10.1145/3555371}{doi:\nolinkurl{10.1145/3555371}}


\end{thebibliography}

\end{document}